\documentclass[11pt]{article}
\usepackage{makecell}
\usepackage[most]{tcolorbox}
\usepackage{hyperref}
\usepackage{url}
\usepackage{multirow}
\usepackage{booktabs}
\usepackage{siunitx}
\usepackage{ulem}
\usepackage{enumitem}
\usepackage{caption}
\usepackage{graphicx}
\usepackage{subcaption}
\usepackage{xcolor}
\usepackage{quoting}
\usepackage{booktabs}
\usepackage{amsmath} 
\usepackage{amsfonts}
\usepackage{xcolor} 
\usepackage{caption} 
\usepackage{float} 
\usepackage{stfloats}
\usepackage{hyperref}
\usepackage{wrapfig}

\usepackage{booktabs}
\usepackage{multirow}
\usepackage{pifont}
\usepackage[table]{xcolor}

\definecolor{highlightcolor}{gray}{0.95}

\tcbuselibrary{breakable}
\usepackage[svgnames]{xcolor}
\usepackage{textcase} 
\tcbset{
  prompt/.style={
    colback=OldLace,
    colframe=YellowGreen,
    width=\linewidth,
    arc=2mm,
    auto outer arc,
    breakable,
    lefttitle=0pt,      
    boxsep=1.5mm,       
    left=1.5mm, right=1.5mm, top=1.5mm, bottom=1.5mm,
    fonttitle=\sffamily,  
    fontupper=\ttfamily,     
    title={\MakeTextUppercase{#1}},
  }
}

\usepackage[utf8]{inputenc}
\usepackage[a4paper, margin=1in]{geometry} 
\usepackage{longtable} 
\usepackage{multirow}  
\usepackage{array}     
\usepackage[table]{xcolor} 
\usepackage{pifont}
\setlist[itemize]{leftmargin=*}
\definecolor{lightgray}{gray}{0.92}
\newcolumntype{C}[1]{>{\centering\arraybackslash}m{#1}}

\usepackage[final]{acl}

\usepackage{times}
\usepackage{latexsym}
\usepackage{indentfirst}
\usepackage[T1]{fontenc}

\usepackage[utf8]{inputenc}

\usepackage{microtype}

\usepackage{inconsolata}

\usepackage{graphicx}

\title{Can LLMs See Without Pixels? Benchmarking Spatial Intelligence from Textual Descriptions}

\author{
 \textbf{Zhongbin Guo\textsuperscript{1}\thanks{Equal contribution.}},
 \textbf{Zhen Yang\textsuperscript{1}\footnotemark[1]},
 \textbf{Yushan Li\textsuperscript{1}},
 \textbf{Xinyue Zhang\textsuperscript{1}},
 \textbf{Wenyu Gao\textsuperscript{1}},
\\
 \textbf{Jiacheng Wang\textsuperscript{1}},
 \textbf{Chengzhi Li\textsuperscript{1}},
 \textbf{Xiangrui Liu\textsuperscript{2}}, 
 \textbf{Ping Jian\textsuperscript{1}\thanks{Corresponding Author.}},
\\
\\
 \textsuperscript{1}School of Computer Science \& Technology, Beijing Institute of Technology,
 \textsuperscript{2}BUCT,
\\
 \small{
   \textbf{Correspondence:} \href{mailto:pjian@bit.edu.cn}{pjian@bit.edu.cn}
 }
}

\begin{document}
\maketitle
\begin{abstract}
Recent advancements in Spatial Intelligence (SI) have predominantly relied on Vision-Language Models (VLMs), yet a critical question remains: does spatial understanding originate from visual encoders or the fundamental reasoning backbone? Inspired by this question, we introduce \textbf{SiT-Bench}, a novel benchmark designed to evaluate the SI performance of Large Language Models (LLMs) without pixel-level input, comprises over 3,800 expert-annotated items across five primary categories and 17 subtasks, ranging from egocentric navigation and perspective transformation to fine-grained robotic manipulation. By converting single/multi-view scenes into high-fidelity, coordinate-aware textual descriptions, we challenge LLMs to perform symbolic textual reasoning rather than visual pattern matching. Evaluation results of state-of-the-art (SOTA) LLMs reveals that while models achieve proficiency in localized semantic tasks, a significant ``spatial gap" remains in global consistency. Notably, we find that explicit spatial reasoning significantly boosts performance, suggesting that LLMs possess latent world-modeling potential. Our proposed dataset \textbf{SiT-Bench} serves as a foundational resource to foster the development of spatially-grounded LLM backbones for future VLMs and embodied agents.
\end{abstract}

\section{Introduction}

Spatial Intelligence (SI)---the ability to perceive, reason about, and interact with the physical world, is a foundational pillar of Embodied Artificial Intelligence~\cite{linMMSIVideoBenchHolisticBenchmark2025,yangCambrianSSpatialSupersensing2025,yangThinkingSpaceHow2024,yinSpatialMentalModeling2025a,zhengMultimodalSpatialReasoning2025}. Recent breakthroughs in Vision-Language Models (VLMs)~\cite{bai2025qwen2,openaiGPT5,Gemini3ProModelCard,claude-4-5-sonnet} as well as series of spatial-enhanced models~\cite{fanVLM3RVisionLanguageModels2025,wuSpatialMLLMBoostingMLLM2025,guo2025beyond} have significantly advanced the field, enabling robotic agents to perform complex tasks ranging from semantic navigation to delicate object manipulation~\cite{song2025robospatial,team2025gemini}. However, these models are typically evaluated on end-to-end visual benchmarks where the synergy between visual perception and linguistic reasoning is treated as a unified capability~\cite{yangThinkingSpaceHow2024,stogiannidis2025mind,zhang2025from}. This coupling masks a fundamental question: \textbf{Does spatial intelligence truly originate from the internal reasoning backbone, or is it merely an artifact of sophisticated pattern matching within the visual encoder?} 

Understanding this distinction is critical for characterizing the symbolic reasoning capacity of Large Language Models (LLMs). In cognitive science, spatial reasoning is often considered a modal-independent process~\cite{jia2025omnispatial}, humans can construct rich mental maps based solely on linguistic descriptions, such as global perception and mapping~\cite{markostamou2024imagery}. Recent studies on multi-view reasoning further support this LLM backbone centric view: while explicit visual enhancements like view interpolation fail to significantly boost VLM performance, enabling free-form textual reasoning or intermediate cognitive mapping leads to substantial improvements~\cite{yinSpatialMentalModeling2025a}. This suggests that for complex spatial understanding, ``thinking" in structured symbolic language is more effective than ``seeing" more pixels. Meanwhile, research on ``language priors" reveals that many VLMs achieve high scores by exploiting linguistic statistical regularities rather than true visual grounding~\cite{lin2024revisiting}. Consequently, if LLMs are to serve as the foundational reasoning engines for future multi-modal systems, they must possess an intrinsic spatial logic capable of manipulating abstract representations independent of immediate visual input.

To bridge this gap, we introduce \textbf{Spatial-in-Text (SiT-Bench)}, a novel and comprehensive benchmark designed to disentangle spatial cognition from visual perception. By evaluating in a vision-ablated, coordinate-aware textual setting, we challenge LLMs to perform pure symbolic geometric reasoning, rigorously determine whether a model possesses genuine internal world model or is simply relying on superficial patterns. As shown in Table~\ref{tab:benchmark_comparison}, SiT-Bench represents a significant leap in scale and diversity compared to previous attempts at textual spatial evaluation, comprising over 3,800 expert-annotated items across five primary categories and 17 subtasks: from egocentric navigation to multi-view perspective stitching, providing a ceiling of spatial intelligence in the post-VLM era~\cite{chen2025era}. 

\begin{table*}[t]
  \centering
  \small
  \begin{tabular}{l c c c c c c c}
  \toprule
  \textbf{Benchmark} & \textbf{Input} & \textbf{Task} & \textbf{Data} & \textbf{Anno.} &  \textbf{Data} & \textbf{Spatial} \\
   & \textbf{Modality} & \textbf{Categories} & \textbf{Domain} & \textbf{Method} & \textbf{Scale} & \textbf{QAs} \\
  \midrule
  RoboSpatial \cite{song2025robospatial} & V & 4 & Indoor, Tabletop & Template & 1M & 3M \\
  VSI-Bench \cite{yangThinkingSpaceHow2024} & V & 8 & Indoor & Template & 1387 & 5K \\
  Ego3D-Bench \cite{gholami2025spatial} & V & 5 & Outdoor & Template & - & 8.6K \\
  BLINK-Spatial \cite{fu2024blink} & V & 14 & MSCOCO & Manual & 286 & 286 \\
  SpatialEval \cite{wang2024is} & V+T & 4 & Maze,Grid,Real & Template & - & 4.6K \\
  FloorplanQA \cite{rodionov2025floorplanqa} & T & 3 & Floor Plans & Template & 2000 & 16K \\
  RoomSpace \cite{li2024reframing} & T & 3 & Virtual Indoor & Template & 10K & 10K \\ 
  \midrule
  \rowcolor{highlightcolor} 
  
   & &  & Indoor, Outdor &  &  &  \\
   \textbf{SiT-Bench (Ours)} & \textbf{T} & \textbf{17} & \textbf{Embodied}, \textbf{Gaming} & \textbf{Manual} & \textbf{2.6K} & 3.9K \\
   & &  & FloorPlan, Tabletop &  &  &  \\
  \bottomrule
  \end{tabular}
\caption{Comparison of \textbf{SiT-Bench} with existing spatial reasoning benchmarks. Our benchmark is the first to provide a large-scale, high-fidelity textual environment that fully decouples spatial cognition from visual perception across the most diverse set of subtasks.}
  \label{tab:benchmark_comparison}
  \vspace{-1.5em}
\end{table*}

Our extensive evaluation of state-of-the-art (SOTA) LLMs reveals nuanced landscape of current spatial capabilities. While modern LLMs demonstrate proficiency in localized semantic tasks, such as identifying immediate neighbor relations, they exhibit a profound ``spatial gap" when challenged with global consistency and complex coordinate transformations. Crucially, our findings indicate that the explicit spatial reasoning (e.g., Chain-of-Thought (CoT)~\cite{wei2022chain}) significantly enhances model performance, suggesting that LLMs possess a notable potential for world modeling that remains underutilized in vanilla prompting.

\textbf{The main contributions of this work can be summarized as follows: }
\vspace{-0.2em}
\begin{itemize}[leftmargin=*]
  \item We introduce \textbf{SiT-Bench}, a large-scale, high-fidelity textual benchmark comprising over 3,800 tailored questions across 5 primary categories (including global perception, embodied tasks, etc.) and 17 diverse subtasks, which decouples spatial reasoning from visual perception, providing a systematic quantitative assessment of LLMs' spatial reasoning capabilities.
  \vspace{-0.2em}
  \item We provide a rigorous evaluation of current SOTA LLMs on SiT-Bench, identify key error patterns in pure-text spatial reasoning, providing empirival insights which can help community develop more reliable LLM backbones for VLMs and embodied applications. 
  \vspace{-0.2em}
  \item The findings in this work uncover the key bottlenecks in achieving genuine SI, providing valuable insights for developing advanced models with stronger spatial intelligence.
\end{itemize}

\section{SiT-Bench: A Textual-Spatial Reasoning Benchmark}
\label{sec:sit_bench}
\begin{figure*}[!t]
    \centering
    \includegraphics[width=0.99\textwidth]{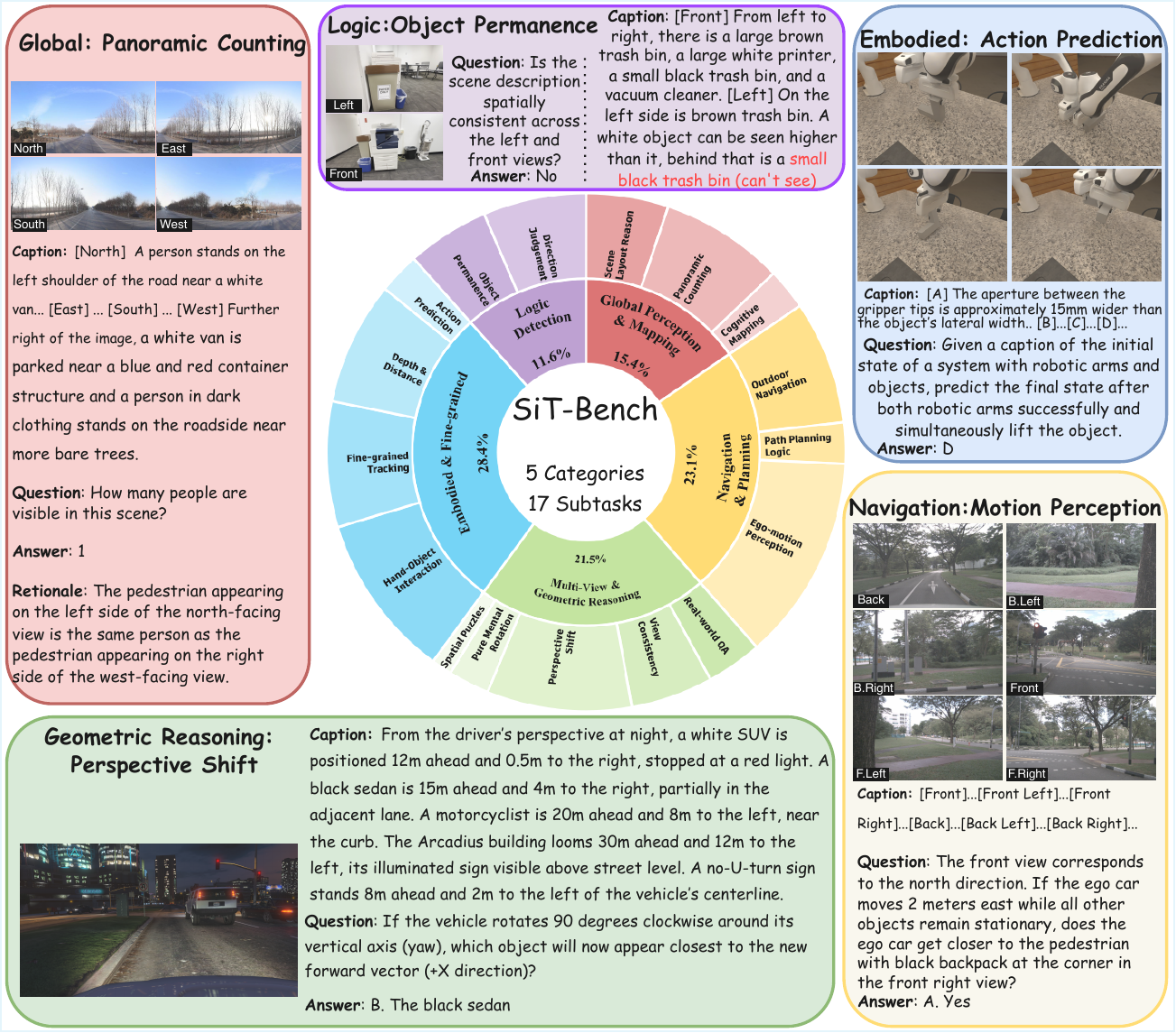}
    \caption{
        \textbf{Tasks Demonstration of SiT-Bench}. Several representative subtasks are selected for demonstration in each of task categories. Note: The images shown are for illustrative purposes only to aid understanding; the actual evaluation uses only textual input without any visual data. The questions and captions above are slightly simplified for clarity and conciseness.
    }
    \label{fig:task_sample}
    \vspace{-0.9em}
\end{figure*}

In this section, we present detailed tasks design and construction pipeline of SiT-Bench. The task samples and constrution pipeline is depicted in Figure~\ref{fig:task_sample} and~\ref{fig:pipeline}. 

\subsection{Task Taxonomy and Design}
\label{subsec:taxonomy}
\noindent We Propose 17 tasks in total, each targeting a distinct facet of spatial cognition, divided these tasks into 5 primary dimensions: \textit{Global Perception \& Mapping}, \textit{Navigation \& Planning}, \textit{Multi-View \& Geometric Reasoning}, \textit{Embodied \& Fine-grained Perception} and \textit{Logic \& Anomaly Detection}.

\noindent\textbf{Global Perception \& Mapping} \\
This dimension evaluates the model’s ability to synthesize fragmented and ego-centric textual cues into a coherent global "mental map". It requires integrating information across wide-angle views or sequential observations to perform \textit{Panoramic Counting} and \textit{Scene Layout Reasoning}. A key innovation is the \textit{Cognitive Mapping} task, inspired by VSI-Bench~\cite{yangThinkingSpaceHow2024} and MindCube~\cite{yinSpatialMentalModeling2025a}, which challenges models to reconstruct unstructured textual navigation logs into structured spatial representations, such as 2D grid layouts or JSON-formatted topological maps.

\noindent\textbf{Navigation \& Planning} \\
Focusing on egocentric decision-making, this category probes the model's capacity for dynamic orientation and long-horizon pathfinding. Utilizing high-fidelity simulated street-view and indoor data, models must execute \textit{Outdoor Navigation} by predicting view changes after specific maneuvers. Furthermore, it assesses \textit{Path Planning Logic} and \textit{Motion Perception}, requiring the model to infer movement vectors (e.g., ego-car displacement) and maintain spatial awareness under continuous coordinate shifts.

\noindent\textbf{Multi-View \& Geometric Reasoning} \\
As the core module of SiT-Bench, this dimension necessitates rigorous 3D geometric modeling and coordinate transformations. Tasks transcend simple semantic matching by requiring \textit{Perspective Shifts} (reasoning from a non-observer POV) and \textit{Pure Mental Rotation} of abstract coordinates. Additionally, it incorporates \textit{Spatial Puzzles} (e.g., LEGO assembly) and \textit{View Consistency} tests to evaluate the understanding of part-whole topological relationships and rotation invariance across arbitrary vertical and horizontal axes.

\noindent\textbf{Embodied \& Fine-grained Perception} \\
This category bridges the gap between abstract reasoning and physical interaction by testing sensitivity to micro-spatial relationships and contact physics. It encompasses \textit{Hand-Object Interaction} geometry, \textit{Relative Depth \& Distance} estimation, and \textit{Fine-grained State Tracking}. Critically, the \textit{Action Prediction} task requires models to predict the physical outcomes of robotic interventions (e.g., the success of a dual-arm lift) based on precise spatial configurations and gripper kinematics.

\noindent\textbf{Logic \& Anomaly Detection} \\
To verify the internal consistency of the model's world model, this dimension assesses adherence to fundamental spatial axioms. Through \textit{Object Permanence} challenges, models must identify logical contradictions or "hallucinated" entities across disjoint viewpoints. Coupled with \textit{Direction Judgement} involving cardinal orientations, we ensures model’s spatial reasoning is grounded in physical reality rather than mere linguistic probability.

\subsection{Benchmark Construction Pipeline}
\label{subsec:pipeline}
We propose a robust, multi-stage pipeline to ensure high data quality and eliminate modality-specific biases (see Figure~\ref{fig:pipeline}). The construction is divided into two parallel paths:

\begin{figure*}[!t]
    \centering
    \includegraphics[width=0.99\textwidth]{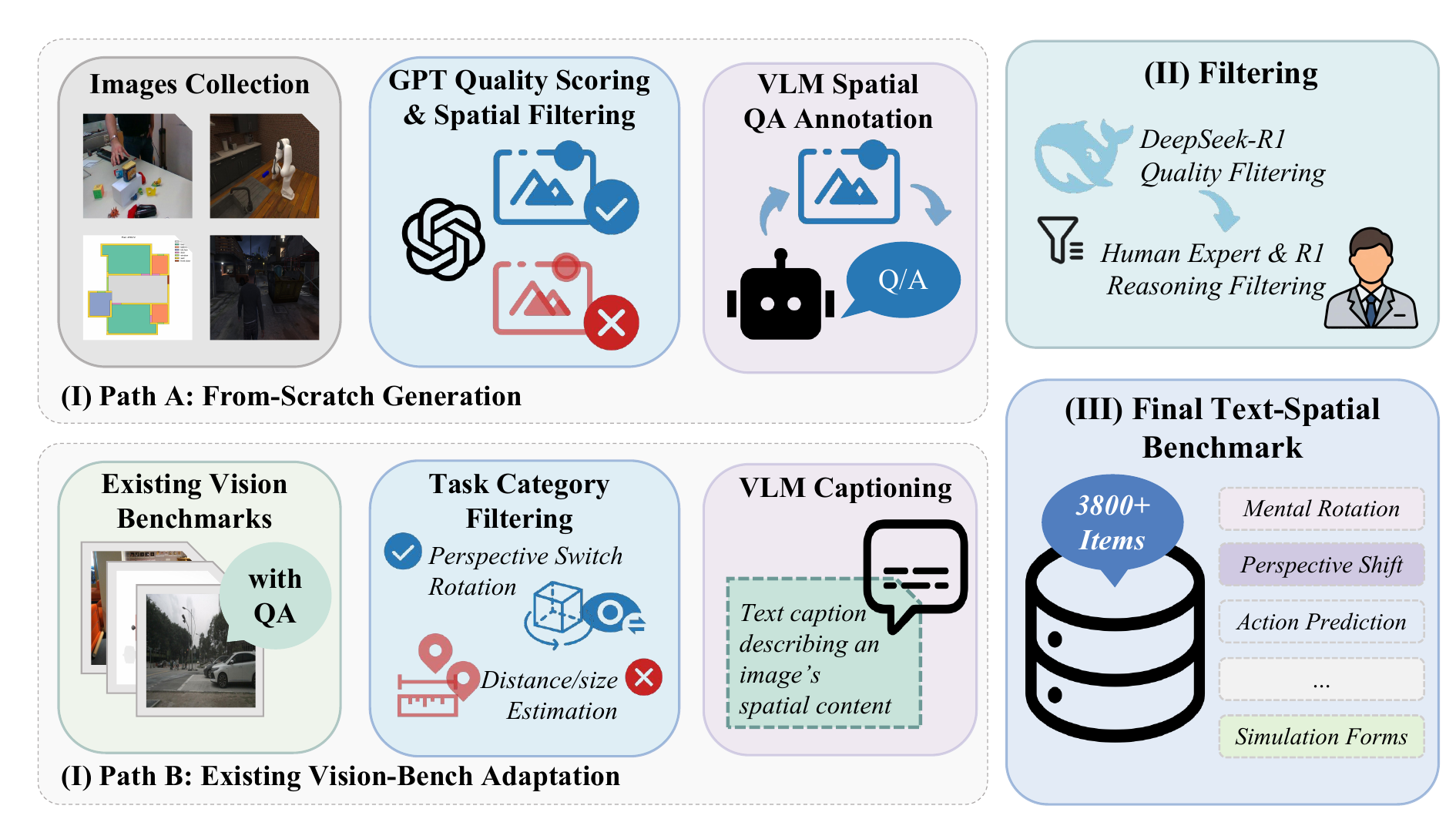}
    \caption{
        \textbf{Benchmark curation pipeline.} The pipeline consists of two parallel paths: \textit{Path A} generates QA pairs from scratch by collecting diverse scene images (robotic manipulation, urban streets, indoor spaces, simulations), applying GPT-4o quality scoring to filter spatially complex samples, and guiding VLMs to produce spatial QA pairs. \textit{Path B} adapts existing vision benchmarks by selecting tasks solvable via pure text (e.g., multi-view reasoning, orientation), captioning their images, and filtering out tasks requiring absolute metrics. Both paths undergo DeepSeek-R1 automated filtering to eliminate data leakage (e.g., direct counting) and caption-uninferrable questions, followed by expert review with R1-CoT rationales to finalize 3,800 high-quality samples.
    }
    \label{fig:pipeline}
    \vspace{-1.2em}
\end{figure*}

\paragraph{Path A: From-Scratch Generation.} 
We collect a diverse set of raw images spanning four major domains: \textit{Robotic Manipulation}~\cite{yang2025multi}, \textit{Residential Floor Plans}~\cite{abouagour2025resplan}, \textit{Open-World Game Scenes}~\cite{Richter_2016_ECCV}, and \textit{Simulated Environments}~\cite{gao2025vision}. We first employ GPT-4o to perform Spatial Quality Scoring, filtering out images with low complexity or insufficient spatial depth. Remaining high-quality images are used to guide VLMs~\cite{bai2025qwen2} in generating location/direction-aware captions and corresponding spatial QA pairs.

\paragraph{Path B: Vision-Bench Adaptation.} 
To compare LLM reasoning directly with visual perception, we select established vision-based benchmarks (e.g., CoSpace~\cite{zhu2025cospace}, ViewSpatial-Bench~\cite{li2025viewspatial} and Ego3D-Bench~\cite{gholami2025spatial}). We carefully filter task types that can be solved through pure textual reasoning—such as multi-view counting, orientation perception, and relative spatial relationships, while excluding tasks that require absolute visual measurements (e.g., absolute distance or object size estimation). We then convert the visual evidence into dense, symbolic textual descriptions, which allows us to assess the ``spatial gap" between LLM backbones and their VLM counterparts on identical logical tasks.

\paragraph{Quality Control and Reasoning-Aware Filtering.} 
Both paths converge into a rigorous two-phase verification process:
\begin{itemize}
    \item \textbf{Phase 1: DeepSeek-R1 Automated Filtering.} We leverage DeepSeek-R1's~\cite{guoDeepSeekR1IncentivizesReasoning2025} advanced reasoning capabilities to perform systematic quality control. For each candidate QA pair, R1 generates a detailed justification explaining whether the sample should be retained or discarded based on multiple filtering criteria, including data leakage detection (e.g., answers derivable through trivial counting or keyword matching, which directly appears in scene captions), caption sufficiency (ensuring all required information is explicitly present without visual hallucination), logical consistency with geometric axioms, and reasoning depth requirements. The complete filtering protocol and R1 prompt templates are provided in Appendix~\ref{app:filtering_protocol}.
    \item \textbf{Phase 2: Human Experts \& R1-CoT Review.} Finally, human experts, assisted by R1's CoT rationales, review each item. We analyze whether the reasoning process is logically sound and geometrically valid. This human-in-the-loop approach ensures the final 3,800+ samples are of professional-grade accuracy.
\end{itemize}

\subsection{Dataset Statistics and Diversity}
\label{subsec:statistics}


SiT-Bench comprises 3,892 samples distributed across five major categories and 17 fine-grained subtasks (see Figure~\ref{fig:task_sample}). The distribution is as follows: \textit{Navigation \& Planning} (23.1\%, 900 samples) focuses on outdoor navigation, ego/object-motion perception, and path planning logic; \textit{Embodied \& Fine-grained} (28.4\%, 1,105 samples) covers hand-object interaction, robotic action prediction, depth/distance perception, and fine-grained state tracking; \textit{Multi-View \& Geometric Reasoning} (21.5\%, 836 samples) includes real-world QA, spatial puzzles, pure mental rotation, view consistency, and perspective shift tasks; \textit{Global Perception \& Mapping} (15.4\%, 601 samples) evaluates panoramic counting, scene layout reasoning, and cognitive mapping; and \textit{Logic Detection} (11.6\%, 450 samples) tests direction judgement and object permanence. By providing explicit distances (m), angular offsets ($^\circ$), and egocentric orientations, SiT-Bench serves as a reproducible and challenging testbed for the next generation of spatially-grounded LLMs.

\section{Experiments}

\subsection{Implementation Details}
We conduct a comprehensive evaluation of state-of-the-art LLMs and VLMs across diverse model families, ranging from compact 3B-parameter models to large-scale models with hundreds of billions of parameters, to assess their spatial reasoning capabilities on SiT-Bench. Our evaluation encompasses both proprietary and open-source solutions.
For proprietary models and large-scale models~(more than 100B params), we evaluate GPT-4o~\cite{hurst2024gpt}, Gemini-3.0-Flash~\cite{gemini3flash} and DeepSeek-V3.2~\cite{liu2025deepseek}. For open-source models, we assess leading VLMs including Qwen2.5/3-VL~\cite{bai2025qwen2}, InternVL3~\cite{zhu2025internvl3}, InternVL3.5~\cite{wang2025internvl3} and LLaVA-1.5~\cite{liu2024improved}, as well as their corresponding LLM backbones (Qwen2.5/3~\cite{yang2024qwen2} and Llama3.1~\cite{grattafiori2024llama}) to directly compare visual perception with pure textual reasoning capabilities. For model series which have reasoning abilities, we evaluate both their non-thinking mode and thinking mode to investigate whether chain-of-thought reasoning can improve spatial reasoning from visual inputs. 
Additional evaluations of larger parameter variants within these model families are provided in Appendix~\ref{app:more_results}.

We include models specifically designed for spatial reasoning: Space-Qwen and SpaceThinker~\cite{chen2024spatialvlm}, Robobrain2.0~\cite{team2025robobrain}, SpaceR~\cite{ouyang2025spacer}, and Cosmos-Reason2~\cite{azzolini2025cosmos}. These models serve as important baselines to understand whether domain-specific architectural designs, training strategies or specific training datasets provide advantages over general-purpose models.

\paragraph{Evaluation Protocol.}
Most tasks in SiT-Bench are presented in a multiple-choice format, while the Cognitive Map subtask requires structured json output matching. We provide random choice baseline scores in our evaluation tables for reference. For multiple-choice tasks, we measure each model's accuracy by directly comparing the model's selected answer with the ground truth. For models with thinking modes, we allow them to generate reasoning traces before producing the final answer, following their default inference protocols.
Concrete implement parameters are shown in Appendix~\ref{app:implement_details}.


\subsection{Main Results}
\label{subsec:main_results}

As shown in Table~\ref{tab:results_categories}, we present a comprehensive evaluation of various LLMs/VLMs on SiT-Bench. The results reveal a substantial gap between human-level performance and current state-of-the-art models, underscoring the challenging nature of our benchmark for spatial reasoning.

\renewcommand{\arraystretch}{1.25} 
\setlength{\tabcolsep}{3pt} 
\definecolor{lightyellow}{RGB}{255, 250, 194}
\definecolor{lightred}{RGB}{250, 189, 164}
\definecolor{lightorange}{RGB}{254, 226, 205}
\begin{table*}[ht]
\footnotesize
    \centering
    \small
    \resizebox{1.0\textwidth}{!}{
    \begin{tabular}{l| c c| c c c c| c c c c| c c c c c c| c c c c c| c c c}
        \textbf{Models} & \rotatebox{75}{\textbf{Rank}} & \normalsize\rotatebox{75}{\textbf{Avg.}} & \normalsize\rotatebox{75}{\makecell{\textbf{Global Perception} \\ \textbf{\& Mapping}}} & 
       \scriptsize \rotatebox{75}{Scene Layout Reason} & \scriptsize \rotatebox{75}{Panoramic Counting} & \scriptsize \rotatebox{75}{Cognitive Mapping} & \normalsize\rotatebox{75}{\makecell{\textbf{Navigation} \\ \textbf{\& Planning}}} & \scriptsize \rotatebox{75}{Outdoor Navigation} & \scriptsize \rotatebox{75}{Path Planning Logic} & \scriptsize \rotatebox{75}{Ego/Objects-motion Perception} & \normalsize\rotatebox{75}{\makecell{\textbf{Multi-View \&} \\ \textbf{Geometric Reasoning}}} & \scriptsize \rotatebox{75}{Real-world QA} & \scriptsize \rotatebox{75}{View Consistency} & \scriptsize \rotatebox{75}{Perspective Shift} & \scriptsize \rotatebox{75}{Pure Mental Rotation} & \scriptsize \rotatebox{75}{Spatial Puzzles}& \normalsize\rotatebox{75}{\makecell{\textbf{Embodied \&} \\ \textbf{Fine-grained}}} & \scriptsize \rotatebox{75}{Hand-Object Interaction} & \scriptsize \rotatebox{75}{Fine-grained Tracking} & \scriptsize\rotatebox{75}{Depth \& Distance} & \scriptsize \rotatebox{75}{Action Prediction} & \normalsize\rotatebox{75}{\textbf{Logic Detection}} & \scriptsize \rotatebox{75}{Object Peranence} & \scriptsize\rotatebox{75}{Direction Judgement}\\
        \hline
        \textbf{Baseline} & & & & & & & & & & & & & & & & & & & & & & & & \\
        Human Level & 1 & 74.42 & 67.85 & 80.00 & 73.42 & 26.77 & 78.22 & 64.67 & 95.00 & 83.00 & 77.45 & 98.51 & 75.00 & 71.23 & 68.00 & 93.00 & 71.86 & 71.50 & 72.13 & 77.67 & 55.00 & 76.22 & 70.00 & 81.20 \\
        Random Level & 32 & 27.30 & - & 25.00 & 25.00 & - & 34.72 & 12.50 & 25.00 & 50.00 & 24.99 & 24.96 & 25.00 & 25.00 & 25.00 & 25.00 & 24.98 & 24.95 & 25.00 & 25.00 & 25.00 & 25.00 & 25.00 & 25.00 \\
        \hline
        \textbf{Proprietary Models / 100B+ Models} & & & & & & & & & & & & & & & & & & & & & & & & \\
        GPT-4o~\cite{hurst2024gpt} & 6 & \cellcolor{lightyellow} 45.70 & 17.74 & 11.50 & 26.58 & 3.61 & 53.78 & 32.00 & 85.00 & 60.60 & \cellcolor{lightyellow} 54.55 & 91.85 & 39.00 & 51.28 & 37.00 & 74.00 & \cellcolor{lightyellow} 47.78 & 30.00 & 56.07 & 70.67 & 25.00 & 45.33 & 74.00 & 22.40 \\
        DeepSeek-V3.2~\cite{liu2025deepseek} & 22 & 37.06 & 19.68 & 13.50 & 29.24 & 3.30 & 49.89 & 19.67 & 87.00 & 60.60 & 46.65 & 93.33 & 38.00 & 33.05 & 36.00 & 72.00 & 33.67 & 29.50 & 39.34 & 38.00 & 20.00 & 25.11 & 21.50 & 28.00 \\
        ~~~-thinking & 10 & 43.74 & \cellcolor{lightyellow} 22.02 & 16.50 & 32.89 & 0.33 & \cellcolor{lightyellow} 61.22 & 12.00 & 86.00 & 85.80 & 53.71 & 97.78 & 37.00 & 47.29 & 37.00 & 80.00 & 32.76 & 13.25 & 55.08 & 37.33 & 29.00 & \cellcolor{lightyellow} 46.22 & 63.00 & 32.80\\
        Gemini-3-Flash-preview~\cite{gemini3flash} & 2 & \cellcolor{lightred} 59.46 & \cellcolor{lightred} 35.66 & 44.50 & 38.87 & 8.34 & \cellcolor{lightred} 77.11 & 47.00 & 89.00 & 92.80 & \cellcolor{lightred} 68.54 & 96.30 & 50.50 & 72.65 & 45.00 & 84.00 & \cellcolor{lightred} 51.31 & 27.75 & 65.25 & 76.67 & 27.00 & \cellcolor{lightred} 59.11 & 61.00 & 57.60 \\
        \hline
        \textbf{Open-Source Models / 100B- Models} & & & & & & & & & & & & & & & & & & & & & & & & \\
        LlaVA-1.5-7B~\cite{liu2024improved} & 31 & 30.53 & \cellcolor{lightred} 29.18 & 28.00 & 39.53 & 0.34 & 39.33 & 16.33 & 95.00 & 42.00 & 29.78 & 28.89 & 31.00 & 31.91 & 25.00 & 22.00 & 25.52 & 23.25 & 30.16 & 22.33 & 30.00 & 28.44 & 22.50 & 33.20 \\
        Llama-3.1-8B~\cite{grattafiori2024llama} & 27 & 34.78 & 14.28 & 15.00 & 17.94 & 1.82 & 45.11 & 17.00 & 71.00 & 56.80 & 36.60 & 88.15 & 31.50 & 21.94 & 27.00 & 40.00 & 39.73 & 51.75 & 34.43 & 31.33 & 33.00 & 26.00 & 19.50 & 31.20 \\
        InternVL3-2B~\cite{zhu2025internvl3} & 29 & 33.92 & 20.68 & 16.50 & 29.90 & 1.28 & 42.67 & 18.00 & 87.00 & 48.60 & 39.59 & 87.41 & 32.00 & 30.48 & 24.00 & 36.00 & 31.22 & 6.75 & 36.39 & 24.67 & 13.00 & 30.22 & 22.50 & 36.40 \\
        InternVL3-8B~\cite{zhu2025internvl3} & 20 & 38.42 & 22.68 & 13.50 & 35.88 & 1.29 & 35.00 & 10.33 & 71.00 & 42.60 & 46.41 & 93.33 & 33.50 & 38.18 & 27.00 & 68.00 & 47.06 & 53.75 & 46.56 & 43.33 & 33.00 & 30.22 & 29.00 & 31.20 \\
        InternVL3.5-4B~\cite{wang2025internvl3} & 17 & 39.95 & 25.79 & 15.00 & 40.20 & 4.01 & 47.44 & 17.67 & 88.00 & 57.20 & 44.50 & 95.56 & 35.50 & 32.48 & 28.00 & 60.00 & 38.73 & 36.50 & 43.93 & 38.00 & 34.00 & 38.44 & 46.50 & 32.00 \\
        ~~~-thinking & 18 & 38.98 & 22.14 & 17.50 & 32.23 & 1.04 & 47.00 & 21.33 & 77.00 & 56.40 & 40.43 & 93.33 & 30.00 & 27.92 & 23.00 & 62.00 & 38.10 & 36.25 & 48.52 & 30.33 & 37.00 & 44.89 & 57.00 & 35.20 \\
        InternVL3.5-8B~\cite{wang2025internvl3} & 12 & 43.27 & 26.14 & 18.50 & 38.87 & 3.09 & 49.78 & 19.33 & 90.00 & 60.00 & 44.26 & 94.81 & 34.50 & 33.05 & 26.00 & 62.00 & 48.78 & 48.00 & 52.46 & 49.33 & 39.00 & 37.78 & 45.50 & 31.60 \\
        ~~~-thinking & 4 & \cellcolor{lightyellow} 46.43 & 18.65 & 14.50 & 27.24 & 1.07 & \cellcolor{lightyellow} 62.00 & 24.33 & 85.00 & 80.00 & 52.87 & 96.30 & 39.00 & 46.72 & 28.00 & 84.00 & 45.61 & 42.75 & 57.05 & 44.00 & 27.00 & 42.44 & 52.50 & 34.40 \\
        Qwen2.5-3B~\cite{yang2024qwen2} & 26 & 34.81 & \cellcolor{lightyellow} 27.93 & 19.50 & 42.52 & 0.83 & 45.44 & 15.67 & 75.00 & 57.40 & 35.05 & 83.70 & 32.50 & 19.66 & 32.00 & 28.00 & 32.85 & 29.25 & 38.03 & 36.00 & 22.00 & 27.11 & 18.00 & 34.40 \\
        Qwen2.5-72B~\cite{yang2024qwen2} & 13 & 42.57 & 14.28 & 15.00 & 17.94 & 1.84 & 50.56 & 26.33 & 90.00 & 57.20 & 53.23 & 95.56 & 39.50 & 48.43 & 35.00 & 64.00 & 47.15 & 36.75 & 43.61 & 70.00 & 31.00 & 33.33 & 32.00 & 34.40 \\
        Qwen2.5-VL-3B~\cite{bai2025qwen2} & 25 & 35.54 & 21.49 & 10.00 & 35.22 & 3.17 & 40.89 & 11.33 & 79.00 & 51.00 & 40.55 & 91.85 & 36.50 & 27.07 & 27.00 & 40.00 & 39.10 & 48.00 & 33.44 & 34.00 & 36.00 & 25.56 & 18.00 & 31.60 \\
        Qwen2.5-VL-72B~\cite{bai2025qwen2} & 8 & 45.45 & 19.29 & 13.00 & 28.90 & 2.94 & 55.67 & 33.33 & 89.00 & 62.40 & \cellcolor{lightyellow} 53.47 & 95.56 & 36.50 & 48.43 & 38.00 & 74.00 & \cellcolor{lightyellow} 49.59 & 33.00 & 52.79 & 76.00 & 27.00 & 34.89 & 47.50 & 24.80 \\
        Qwen3-4B~\cite{yang2024qwen2} & 28 & 34.68 & 12.44 & 16.00 & 13.29 & 2.76 &  45.89 & 20.33 & 84.00 & 53.60 & 43.06 & 87.41 & 37.50 & 34.76 & 25.00 & 40.00 & 34.57 & 38.25 & 36.07 & 29.67 & 30.00 & 26.67 & 20.00 & 32.00 \\
        ~~~-thinking & 14 & 42.26 & 17.24 & 13.00 & 25.25 & 1.62 & 52.67 & 22.67 & 75.00 & 66.20 & \cellcolor{lightyellow} 53.47 & 91.11 & 41.00 & 50.71 & 25.00 & 78.00 & 39.73 & 33.50 & 44.59 & 48.00 & 25.00 & 40.22 & 48.50 & 33.60 \\
        Qwen3-8B~\cite{yang2024qwen2} & 21 & 37.91 & 18.20 & 14.00 & 26.58 & 1.40 & 45.11 & 19.33 & 66.00 & 56.40 & 41.87 & 91.11 & 32.00 & 31.62 & 24.00 & 56.00 & 42.99 & 43.75 & 37.70 & 48.67 & 39.00 & 30.00 & 26.50 & 32.80 \\
        ~~~-thinking & 9 & 45.04 & 17.49 & 13.50 & 25.58 & 1.13 & 58.78 & 22.00 & 72.00 & 78.20 & 52.51 & 94.07 & 34.50 & 49.57 & 27.00 & 84.00 & 44.16 & 39.75 & 47.87 & 51.67 & 28.00 & 42.67 & 48.00 & 38.40 \\
        
        Qwen3-VL-4B~\cite{bai2025qwen2} & 19 & 38.67 & 18.81 & 12.50 & 28.57 & 2.04 & 47.44 & 17.00 & 86.00 & 58.00 & 45.81 & 94.07 & 34.50 & 37.32 & 26.00 & 60.00 & 38.19 & 37.00 & 37.38 & 42.00 & 34.00 & 35.56 & 34.00 & 36.80 \\
        ~~~-thinking & 11 & 43.70 & 15.81 & 15.50 & 21.26 & 0.00 & 58.00 & 22.00 & 80.00 & 75.20 & 51.79 & 92.59 & 37.50 & 46.72 & 28.00 & 82.00 & 39.91 & 29.00 & 44.26 & 55.67 & 23.00 & \cellcolor{lightyellow} 46.67 & 54.00 & 40.80 \\
        Qwen3-VL-8B~\cite{bai2025qwen2} & 16 & 42.10 & 25.74 & 11.50 & 43.52 & 0.69 & 45.78 & 20.67 & 81.00 & 53.80 & 48.44 & 92.59 & 28.50 & 47.01 & 24.00 & 68.00 & 43.53 & 41.75 & 43.28 & 51.00 & 29.00 & 41.33 & 45.00 & 38.40 \\
        ~~~-thinking & 7 & 45.66 & 20.97 & 16.00 & 31.23 & 0.00 & 59.11 & 27.00 & 77.00 & 74.80 & 52.99 & 94.81 & 37.50 & 52.14 & 28.00 & 58.00 & 43.62 & 30.50 & 49.84 & 60.00 & 28.00 & 43.11 & 51.50 & 36.40 \\
        Qwen3-VL-32B~\cite{bai2025qwen2} & 5 & 45.90 & 15.74 & 12.00 & 22.92 & 1.61 & 59.44 & 31.67 & 87.00 & 70.60 & 45.81 & 98.52 & 35.50 & 30.77 & 39.00 & 64.00 & 53.67 & 45.25 & 54.75 & 72.67 & 27.00 & 40.22 & 42.50 & 38.40 \\
        ~~~-thinking & 3 & \cellcolor{lightred} 51.06 & 16.34 & 13.00 & 23.92 & 0.20 & \cellcolor{lightred} 68.67 & 28.67 & 77.00 & 91.00 & \cellcolor{lightred} 59.45 & 96.30 & 39.00 & 59.54 & 40.00 & 80.00 & \cellcolor{lightred} 49.68 & 33.50 & 58.69 & 69.00 & 29.00 & \cellcolor{lightred} 50.00 & 54.00 & 46.80 \\

        \hline
        \textbf{Spatial Models} & & & & & & & & & & & & & & & & & & & & & & & & \\   
        Space-Qwen-3B~\cite{chen2024spatialvlm} & 33 & 27.26 & 16.35 & 21.00 & 17.28 & 4.24 & 36.33 & 11.33 & 71.00 & 44.40 & 27.75 & 44.44 & 22.50 & 25.64 & 24.00 & 26.00 & 29.77 & 27.00 & 38.36 & 28.85 & 18.00 & 16.22 & 19.00 & 14.00 \\
        SpaceThinker-3B~\cite{chen2024spatialvlm} & 30 & 33.83 & \cellcolor{lightred} 20.73 & 18.50 & 28.24 & 2.58 & 43.11 & 12.00 & 61.00 & 58.20 & 38.04 & 86.67 & 36.00 & 25.36 & 21.00 & 38.00 & 32.22 & 32.00 & 32.13 & 33.33 & 30.00 & \cellcolor{lightyellow} 28.89 & 16.00 & 39.20 \\
        Robobrain2.0-7B~\cite{team2025robobrain} & 24 & 35.52 & 18.41 & 16.50 & 25.58 & 0.62 & 36.78 & 17.67 & 67.00 & 42.20 & \cellcolor{lightyellow} 46.17 & 92.59 & 33.50 & 39.32 & 26.00 & 60.00 & \cellcolor{lightyellow} 41.36 & 40.50 & 40.66 & 46.67 & 31.00 & 21.78 & 23.00 & 20.80 \\
        SpaceR-7B~\cite{ouyang2025spacer} & 23 & \cellcolor{lightyellow} 36.42 & 19.40 & 12.50 & 27.91 & 7.60 & \cellcolor{lightyellow} 44.22 & 13.33 & 72.00 & 57.20 & 43.90 & 93.33 & 36.50 & 29.91 & 33.00 & 60.00 & 37.56 & 37.75 & 44.59 & 32.67 & 30.00 & 26.89 & 31.50 & 23.20 \\
        Cosmos-Reason2-8B~\cite{azzolini2025cosmos} & 15 & \cellcolor{lightred} 42.13 & \cellcolor{lightyellow} 20.59 & 14.50 & 31.23 & 0.76 & \cellcolor{lightred} 47.89 & 21.00 & 89.00 & 55.80 & \cellcolor{lightred} 50.00 & 92.59 & 33.00 & 49.86 & 23.00 & 58.00 & \cellcolor{lightred} 43.98 & 37.50 & 44.26 & 56.67 & 31.00 & \cellcolor{lightred} 40.22 & 49.00 & 33.20 \\
    \end{tabular}
    }
    \caption{\textbf{Performance of different models on SiT bench.} The highest and second-highest in each category are highlighted with \colorbox{lightred}{light red} and \colorbox{lightyellow} {light yellow}, respectively.}
    \label{tab:results_categories}
    \vspace{-5mm}
\end{table*}

\paragraph{Overall Performance and the Human Gap.} 
Among all evaluated models, Gemini-3-Flash achieves the strongest performance of 59.46\%, significantly outperforming other proprietary and open-source alternatives. Qwen3-VL-32B-thinking follows with an average accuracy of 51.06\%, leads among open-source models. Despite these strong results, a significant gap remains compared to the \textbf{Human Level (74.42\%)}. Notably, while humans excel in tasks requiring global consistency like \textit{Panoramic Counting} (73.42\%) and \textit{Outdoor Navigation} (64.67\%), even the best-performing models struggle to achieve 10\% accuracy in \textit{Cognitive Mapping} (best: Gemini thinking at 8.34\%, vs. Human at 26.44\%), suggesting that high-level topological reconstruction remains a formidable challenge for current AI.

\paragraph{Scaling vs. Reasoning Backbone.} 
Analysis across model scales indicates that while parameter scaling generally improves performance (e.g., Qwen2.5-3B at 34.81\% vs. Qwen2.5-72B 42.57\%), it is not the sole determinant of spatial intelligence. Notably, almost all reasoning-enabled models exhibit significant performance gains when thinking mode is activated. For example, Qwen3-VL-32B improves from 45.9\% to 51.06\%, and Qwen3-8B jumps from 37.91\% to 45.04\% with thinking enabled. More strikingly, smaller models with thinking capabilities can surpass much larger models without explicit reasoning: the 32B Qwen3-VL-thinking (51.06\%) significantly outperforms the much larger DeepSeek-V3.2 (37.06\%), despite the latter having substantially more parameters. This indicates that explicit chain-of-thought reasoning is more effective than brute-force scaling for complex spatial reasoning tasks.

\paragraph{Performance Across Task Categories.} 
A clear hierarchy of task difficulty emerges:
\begin{itemize}[leftmargin=*]
    \item \textbf{High-Level Semantics:} Tasks like \textit{Real-world QA} show highest scores, with models leveraging linguistic priors effectively.
    \item \textbf{Geometric Transformations:} \textit{Perspective Shift} and \textit{Mental Rotation} see a sharp decline, where models must perform explicit coordinate-frame transformations.
    \item \textbf{Global Consistency:} \textit{Cognitive Mapping} and \textit{Panoramic Counting} remain the most difficult, as they require the persistent maintenance of an internal "world model" to resolve entity overlaps across viewpoints.
\end{itemize}

\paragraph{Random Baseline Comparison.} The random baseline achieves 27.3\% average accuracy. While all models exceed this baseline, the margins for challenging subtasks like Cognitive Mapping and Scene Layout Reasoning remain concerningly small, indicating that models may rely on superficial patterns rather than genuine spatial reasoning for these tasks. These results collectively demonstrate that despite rapid progress in multimodal AI, achieving human-level spatial intelligence remains an open challenge requiring fundamental advances in geometric reasoning, mental simulation, and embodied understanding.

\subsection{Experimental Analysis}
\label{subsec:experimental_analysis}

\paragraph{Visual Grounding Enhances Spatial Understanding in LLM Backbones.} 
One finding is that VLMs consistently outperform their pure LLM backbones even in this vision-ablated textual benchmark. For example, Qwen2.5-VL-72B (45.45\%) surpasses Qwen2.5-72B (42.57\%), and Qwen3-VL-8B (42.10\%) outperforms Qwen3-8B (37.91\%). This suggests that exposure to visual information during VLM training helps the LLM backbone develop a better understanding of real-world spatial relationships. The multimodal training process, through seeing millions of images, effectively ``bakes'' spatial priors into the language weights, providing the model with a more grounded "spatial vocabulary"~(eg. relative direction) and improved comprehension of perspective-dependent descriptions—capabilities it can leverage even when visual inputs are removed.

\begin{figure*}[!t]
    \centering
    \includegraphics[width=0.99\textwidth]{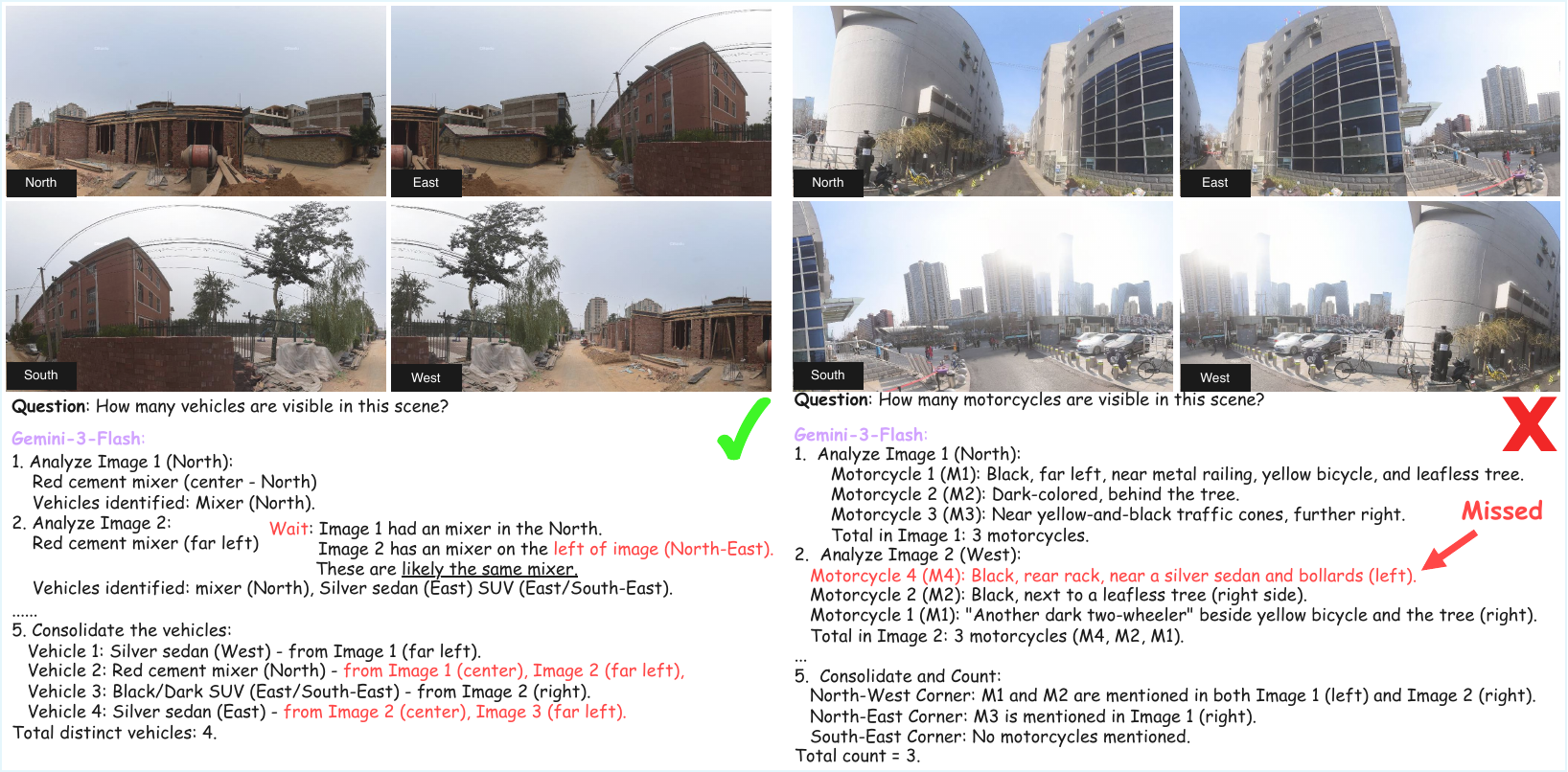}
    \caption{
        The simplified thought process examples of Gemini-3-Flash. Complete reasoning process in Appendix~\ref{app:reasoning_process}
    }
    \label{fig:error_analysis}
    \vspace{-1.2em}
\end{figure*}

\paragraph{The Emergence of Explicit Spatial Reasoning.} 
The transition from "non-thinking" to "thinking" modes provides the most substantial performance leap. As seen in Qwen3-8B, the score significantly increases from 37.91\% to 45.04\% (+7.13\%). Qualitative analysis about the reasoning traces of Gemini-3-Flash in Fig~\ref{fig:error_analysis} reveals that \textbf{reasoning-enabled models explicitly simulate spatial axioms}. For example, in \textit{Panoramic Counting}, a correct Gemini-thinking trace explicitly noted: \textit{"Mixer at middle of North View maybe the same one at left of East View."} Conversely, incorrect traces often fail due to "arithmetic-spatial hallucinations", where the model correctly identifies entity overlaps but miscalculates the final sum or the exact coordinate offset.

\paragraph{Underperformance of Specialized Spatial Models.} 
A surprising result is that specialized spatial models (e.g., Robobrain2.0, SpaceR, SpaceThinker) generally perform worse than, or only comparable to, general-purpose LLMs/VLMs of the same scale. For example, Cosmos-Reason2-8B (42.13\%) is outperformed by the general-purpose Qwen3-VL-8B (45.66\%). This phenomenon contradicts the conventional wisdom that domain-specific fine-tuning is superior. The reason could be that: (i) spatial-specific datasets may be too narrow or template-reliant, causing the model to lose the general reasoning flexibility required for SiT-Bench's diverse scenarios; (ii) spatial training might lead to "catastrophic forgetting" of the broad linguistic common sense needed to parse high-fidelity textual descriptions.

\paragraph{Decoupling Perception and Reasoning.} 
The disparity between VLM performance on SiT-Bench and traditional vision benchmarks provides new insights into the nature of spatial intelligence. High scores on vision benchmarks may be inflated by visual pattern matching. Our benchmark demonstrates that when the reasoning component is isolated and tested independently, even SOTA models encounter performance ceiling. This underscores that current spatial intelligence remains heavily perception-reliant, and building a truly cognitive ``World Model'' requires a fundamental shift toward \textit{internal symbolic manipulation} of spatial representations, a capability that reasoning-augmented models (e.g., Gemini-3-Flash-thinking and Qwen3-thinking) are only beginning to exhibit. These findings suggest that future research should prioritize the development of explicit spatial reasoning mechanisms within language models, rather than relying solely on scaling visual encoders. SiT-Bench provides a principled framework for tracking progress along this dimension, enabling the community to systematically evaluate advances in the cognitive foundations of spatial intelligence.

\section{Related Work}
\label{sec:related_work}


\paragraph{Enhancing Spatial Reasoning in VLMs.} 
Spatial intelligence refers to the cognitive ability to perceive, represent, and reason about spatial relationships, object configurations, and geometric transformations~\cite{yangThinkingSpaceHow2024}. Recent research has made significant strides in improving the spatial capabilities of Vision-Language Models. SpatialRGPT~\cite{chengSpatialRGPTGroundedSpatial2024} and SpatialBot~\cite{caiSpatialBotPreciseSpatial2024} injected depth and region-level 3D features into vision encoders to improve relational judgments. More recently, including MM-Spatial~\cite{daxbergerMMSpatialExploring3D2025} and Video-3D LLM~\cite{zhengVideo3DLLMLearning2025}, integrated 3D reconstruction with video modeling to unify the representation spaces of frames, point clouds, and text. These approaches established an essential foundation for spatio-temporal 3D understanding. Specialized models like Robobrain2.0~\cite{team2025robobrain}, SpaceR~\cite{ouyang2025spacer}, and Cosmos-Reason series~\cite{azzolini2025cosmos} have been developed with explicit spatial training objectives. However, as our experimental analysis reveals, it remains unclear whether these improvements stem from enhanced visual feature extraction or genuine advances in the underlying spatial reasoning of the language backbone.

\paragraph{Benchmarking Spatial Capabilities.} 
Several benchmarks have been proposed to evaluate spatial perception in multimodal models. VSI-Bench~\cite{yangThinkingSpaceHow2024} and View-SpatialBench~\cite{li2025viewspatial} focus on multi-view, video consistency and object localization, while Cambrian-S~\cite{yangCambrianSSpatialSupersensing2025} explores spatial scaling laws of VLMs. However, these benchmarks are intrinsically tied to visual perception, making it difficult to isolate the reasoning component from perceptual pattern matching. Existing text-only spatial evaluations, such as Resplan~\cite{abouagour2025resplan} and FloorplanQA~\cite{rodionov2025floorplanqa} or basic navigation tasks in BigBench~\cite{srivastava2023beyond}, rely on overly simplified 2D grids that fail to capture the complexity of real-world spatial reasoning. To our knowledge, there lacks a comprehensive benchmark that evaluates spatial intelligence purely through high-fidelity textual descriptions. \textbf{SiT-Bench} fills this gap by introducing coordinate-aware 3D descriptions across 17 diverse subtasks, enabling rigorous assessment of the symbolic spatial reasoning capabilities within LLMs.

\section{Conclusions}
In this paper, we introduced \textbf{SiT-Bench}, a large-scale, high-fidelity textual benchmark designed to disentangle spatial cognition from visual perception. By evaluating SOTA models on 3,800+ samples across 17 subtasks, we provided a rigorous assessment of the "reasoning backbone" that powers modern embodied agents. Our results reveal that while LLMs excel at localized spatial semantics, they face substantial challenges in global mental modeling and perspective stitching. However, the marked improvement seen with explicit reasoning suggests that the LLM backbone has untapped potential for world modeling. We believe thSat \textbf{SiT-Bench} will serve as a foundational resource for the community, guiding the development of more spatially-grounded LLMs and facilitating the leap toward truly intelligent embodied agents.

\section{Limitations}
\label{sec:limitations}

Despite the comprehensive nature of \textbf{SiT-Bench}, several limitations remain that offer avenues for future research. 

\paragraph{Discrete Snapshot vs. Continuous Dynamics.} While \textbf{SiT-Bench} covers complex movement through tasks like \textit{Ego-motion Perception} and \textit{Path Planning}, these are grounded in discrete multi-view snapshots or sequential "state-captions". In actual embodied scenarios, spatial intelligence requires processing high-frequency continuous temporal data. Our textual abstraction, while effective for testing topological reasoning, does not fully capture the real-time feedback loops required for low-level motor control in robotics.

\paragraph{Computational Latency of Reasoning Models.} Our experimental analysis highlights that "thinking" modes (e.g., Gemini-3-Flash with CoT) significantly bridge the "spatial gap". However, the substantial computational overhead and latency associated with these explicit reasoning traces currently limit their deployment in latency-sensitive embodied tasks. Bridging the gap between the high-level spatial reasoning observed in our benchmark and the efficiency required for real-time interaction remains an open challenge. We make detailed discussion in Appendix~\ref{app:inference_latency}

\section{Ethics Statement}
\label{sec:ethics_statement}

In the development of \textbf{SiT-Bench}, we have adhered to the highest ethical standards in data collection and model evaluation.

\paragraph{Data Privacy and PII.} All image sources used for caption generation, including those from \textit{GTA-V} (Play4Data) and egocentric datasets (Ego3d), were screened to ensure the absence of Personally Identifiable Information (PII). For urban and indoor scenes, we prioritized simulated or anonymized environments to avoid privacy infringements related to real-world locations or individuals.

\paragraph{Mitigating Demographic and Geographic Bias.} We acknowledge that spatial datasets often reflect geographic biases (e.g., urban structures in Western cities). To mitigate this, \textbf{SiT-Bench} intentionally incorporates a diverse array of 17 subtasks ranging from abstract geometric puzzles and LEGO assembly to various robotic manipulation scenarios. This diversity reduces the reliance on specific cultural or geographic landmarks for spatial reasoning.

\paragraph{Commitment to Open Research.} To foster transparency and reproducibility in the embodied AI community, we commit to releasing the full \textbf{SiT-Bench} dataset, comprising 3,800+ expert-annotated samples. We believe that by open-sourcing these coordinate-aware textual descriptions, we can provide a neutral testbed for assessing the "World Models" of future autonomous agents, thereby preventing the monopolization of spatial intelligence benchmarks by proprietary visual-only platforms.
\bibliography{custom}

\appendix
\twocolumn
\section{Data Sources and Task Distribution}
\label{sec:appendix}

\subsection{Data Sources}

We utilize several publicly available datasets, each contributing specific subsets tailored for spatial reasoning and perception tasks.

\textbf{SpatialEval~\cite{wang2024is}:} We select the "Spatial real" subset, which focuses on real-world spatial commonsense question answering. It tests models on their ability to reason about the relative positioning of objects in a scene, such as determining if one object is to the left or right of another.

\textbf{SpinBench~\cite{zhang2025spinbench}:} We use the "Scene perspective select" subset, which challenges models to reason about different perspectives of the same scene, and the "View-spatial" subset, which tests the consistency of object recognition across different viewpoints. These tasks evaluate the model's ability to handle multiple images depicting the same environment from various angles.

\textbf{Mindcube~\cite{yinSpatialMentalModeling2025}:} For this dataset, we focus on multi-view images of objects and scenes. Specifically, we use samples where the camera rotates around a fixed object, ensuring a variety of perspectives that test models on their ability to recognize spatial consistency across different views.

\textbf{Resplan~\cite{abouagour2025resplan}:} The "Route plan" subset consists of 2D floor plans, typically of room layouts. It tests models on path planning and navigation, specifically how well they can reason about the positioning of rooms and doors to infer possible movements or navigation routes based on structured textual descriptions.

\textbf{Play4Data~\cite{Richter_2016_ECCV}:} This dataset provides two subsets: "Mental Rotation," which evaluates depth and distance perception, and "Perspective shift," which tests the model’s ability to infer the relative positions of objects as the viewpoint changes. These tasks assess spatial reasoning from different angles and distances.

\textbf{POEM-v2~\cite{yang2025multi}:} This dataset includes "Single view" and "Multiview" subsets. The "Single view" subset focuses on hand-object interactions, while "Multiview" examines spatial relationships between objects and the human hand from multiple viewpoints, assessing the model's embodied spatial perception.

\textbf{LEGO-puzzles~\cite{anonymous2025legopuzzles}:} We use this dataset to test models on spatial reasoning involving LEGO structures. The task requires models to determine the perspective of a LEGO structure, testing their ability to infer spatial relationships based on a top-down view.

\textbf{Ego3D-Bench~\cite{gholami2025spatial}:} This dataset offers several subsets, such as "Ego-centric motion" and "Object-centric motion," each involving six viewpoints per data entry. These tasks test models on motion perception, requiring them to infer the movement of objects or agents from different spatial perspectives.

\textbf{Cospace~\cite{zhu2025cospace}:} From this dataset, we select various subsets such as "Angle," which involves object counting and angle measurement, and "Counting," which challenges models to accurately count the number of objects in a scene. Tasks such as "Dif-ang" and "Direction judge" assess the model's ability to determine object permanence and directional orientation in dynamic environments.

\textbf{WM-ABench~\cite{gao2025vision}:} We utilize subsets that focus on object spatial arrangement, geometric object placement, and simulated street view navigation tasks. These tasks involve reasoning about the layout of objects and the potential paths that could be taken in a scene, testing spatial navigation and embodied interaction.

In total, we carefully selected and processed 34,917 samples from these datasets. These samples will be used to construct a benchmark with approximately 4,000 samples and a larger dataset of around 20,000 samples, ensuring a balanced representation of tasks and data diversity.

\subsection{Detailed Task Distribution}

The tasks in our benchmark are categorized into five major categories: \textbf{Global Perception \& Mapping}, \textbf{Navigation \& Planning}, \textbf{Multi-View \& Geometric Reasoning}, \textbf{Embodied \& Fine-grained}, and \textbf{Logic Detection}. Each category is further divided into sub-tasks, as detailed in Table~\ref{tab:category}. This table provides information on the data sources, sample sizes, and task descriptions for each sub-task.

\subsection{Comparison of Model Inference Latency}
\label{app:inference_latency}

As discussed in the main text, while explicit reasoning modes (e.g., Chain-of-Thought prompting) significantly enhance spatial reasoning performance, they introduce substantial computational overhead. Table~\ref{tab:inference_latency} presents a comprehensive comparison of average inference latency across all evaluated models.

\begin{table}[h]
\centering
\small
\begin{tabular}{l r}
\toprule
\textbf{Model Name} & \textbf{Avg Latency (s)} \\
\midrule
\multicolumn{2}{l}{\textit{Standard Inference Models}} \\
\midrule
Cosmos-Reason2-2B & 0.34 \\
InternVL3-2B & 0.27 \\
Llama-3-8B-Instruct & 0.38 \\
llava-1.5-7b-hf & 0.39 \\
Qwen2.5-3B-Instruct & 0.40 \\
Qwen3-4B & 0.41 \\
Qwen2.5-VL-3B-Instruct & 0.42 \\
Qwen3-VL-4B-Instruct & 0.43 \\
Qwen3-4B-Instruct-2507 & 0.39 \\
Cosmos-Reason2-8B & 0.44 \\
InternVL3-8B & 0.46 \\
Qwen3-VL-8B-Instruct & 0.49 \\
InternVL3\_5-30B-A3B & 0.50 \\
Llama-3.1-8B-Instruct & 0.52 \\
Qwen3-8B & 0.52 \\
InternVL3\_5-4B-Instruct & 0.44 \\
InternVL3\_5-8B-Instruct & 0.55 \\
InternVL3\_5-14B-Instruct & 0.55 \\
Qwen3-30B-A3B-Instruct-2507 & 0.59 \\
Qwen2.5-7B-Instruct & 0.62 \\
Qwen2.5-VL-7B-Instruct & 0.65 \\
Qwen3-VL-30B-A3B-Instruct & 0.72 \\
InternVL3-14B & 0.82 \\
gpt-4o & 0.95 \\
Qwen3-VL-32B-Instruct & 1.03 \\
Qwen2.5-VL-72B-Instruct & 1.88 \\
deepseek-v3.2 & 2.71 \\
Qwen2.5-72B-Instruct & 4.94 \\
\midrule
\multicolumn{2}{l}{\textit{Thinking/CoT-Enabled Models}} \\
\midrule
RoboBrain2.0-7B\_thinkon & 0.61 \\
SpaceQwen2.5-VL-3B-Instruct\_thinkon & 2.31 \\
SpaceThinker-Qwen2.5VL-3B\_thinkon & 3.82 \\
SpaceR\_thinkon & 3.87 \\
Qwen3-4B\_thinkon & 14.42 \\
Qwen3-8B\_thinkon & 17.11 \\
gemini-3-flash-preview\_thinkon & 40.11 \\
InternVL3\_5-4B\_thinkon & 43.41 \\
InternVL3\_5\_8B\_thinkon & 46.95 \\
Qwen3-VL-30B-A3B-Thinking\_thinkon & 53.95 \\
InternVL3\_5-38B & 62.57 \\
Qwen3-VL-4B-Thinking\_thinkon & 65.91 \\
Qwen3-VL-32B-Thinking\_thinkon & 73.42 \\
Qwen3-VL-8B-Thinking\_thinkon & 87.99 \\
deepseek-v3.2\_thinkon & 190.24 \\
\bottomrule
\end{tabular}
\caption{Average inference latency comparison across models. Models with ``\_thinkon'' suffix indicate explicit reasoning/thinking mode enabled.}
\label{tab:inference_latency}
\end{table}

\paragraph{Key Observations.} The latency data reveals a stark trade-off between reasoning capability and computational efficiency:

\begin{itemize}[leftmargin=*]
    \item \textbf{Standard models} exhibit sub-second latency in most cases, with lightweight models like Cosmos-Reason2-2B (0.34s) and InternVL3-2B (0.27s) achieving the fastest inference times, making them suitable for real-time applications.
    
    \item \textbf{Thinking-enabled models} demonstrate dramatically increased latency, often by 1-2 orders of magnitude. For instance, Qwen3-4B increases from 0.41s to 14.42s when thinking mode is enabled (35$\times$ increase), while deepseek-v3.2 escalates from 2.71s to 190.24s (70$\times$ increase).
    
    \item \textbf{Spatial-specialized models} such as SpaceQwen2.5-VL-3B-Instruct (2.31s) and SpaceR (3.87s) achieve a reasonable balance, providing enhanced spatial reasoning with moderate latency overhead compared to general thinking models.
    
    \item \textbf{API-based models} like gpt-4o (0.95s) and gemini-3-flash-preview with thinking (40.11s) show that even commercial solutions face significant latency increases when explicit reasoning is required.
\end{itemize}

\paragraph{Implications for Embodied AI.} These findings underscore a critical challenge for deploying spatially-intelligent models in real-world embodied systems. While our benchmark demonstrates that explicit reasoning significantly improves spatial understanding, the associated latency (often exceeding 40-90 seconds per query) is incompatible with the real-time requirements of robotic manipulation, autonomous navigation, and interactive agents. Future research should focus on: (1) distilling spatial reasoning capabilities into more efficient architectures, (2) developing hybrid approaches that selectively engage deep reasoning only when necessary, and (3) exploring hardware acceleration strategies for reasoning-intensive computations.

\begin{table*}[ht]
\footnotesize
\centering
    \resizebox{\textwidth}{!}{ 
    
    \begin{tabular}{ m{3cm}| c| c| c| m{7cm}}
    
    \hline
    \textbf{Major Category} & 
    \textbf{Sub-task} & 
    \textbf{Data Source} & 
    \textbf{Count} & 
    \textbf{Rationale} \\ 
    \hline

    \multirow{4}{=}{\centering \textbf{Global Perception \& Mapping}} 
    & \textbf{Scene Layout Reason} & Mindcube & 200 & Tests the model's ability to reason about common sense spatial relationships. \\ \cline{2-5} 
    & \textbf{Panoramic Counting} & CoSpace & 301 & Assesses the model's ability to count multiple objects and understand spatial relations in a panoramic context. \\ \cline{2-5} 
    & \textbf{Cognitive Mapping} & MindCube & 100 & Evaluates the model’s ability to generate grid layouts or JSON formatted maps from room descriptions. \\ \hline
    
    \multirow{6}{=}{\centering \textbf{Navigation \& Planning}} 
    & \textbf{Outdoor Navigation} & CoSpace / WM-Abench & 300 & Tests the model’s navigation abilities in outdoor environments based on textual descriptions of paths and directions. \\ \cline{2-5} 
    
    & \textbf{Path Planning Logic} & Resplan & 100 & Evaluates the model’s ability to plan routes from point A to B using structured textual descriptions of room layouts. \\ \cline{2-5} 
    
    & \textbf{Ego/Objects-motion Perception} & Ego3d & 500 & Assesses the model's ability to perceive and understand movement directions (forward, back, left, right) from text descriptions. \\ \hline

    \multirow{8}{=}{\centering \textbf{Multi-View \& Geometric Reasoning}} 
    & \textbf{Real-world QA} & SpatialEval & 135 & Tests the model's spatial commonsense knowledge in real-world scenarios to assess its practical understanding of space. \\ \cline{2-5} 
    
    & \textbf{View Consistency} & SpinBench / View-spatial & 200 & Evaluates the model's consistency in recognizing objects from different perspectives within the same scene. \\ \cline{2-5} 
    
    & \textbf{Perspective Shift} & Play4Data / Ego3d & 351 & Assesses the model's ability to infer object positions when the viewpoint changes. \\ \cline{2-5} 
    
    & \textbf{Pure Mental Rotation} & LEGO-puzzles & 100 & Tests the model's ability to perform abstract geometric rotations, removing semantic interference. \\ \cline{2-5} 
    
    & \textbf{Spatial Puzzles} & SpatialEval & 50 & Involves high-difficulty tasks where the model must assemble or disassemble objects based on spatial relations. \\ \hline

    \multirow{8}{=}{\centering \textbf{Embodied \& Fine-grained}} 
    & \textbf{Hand-Object Interaction} & POEM-v2 & 400 & Assesses the model’s ability to understand fine-grained spatial interactions, such as hand-object contact points and relative positions. \\ \cline{2-5} 
    
    & \textbf{Fine-grained Tracking} & WM-Abench & 305 & Tests the model’s ability to track subtle changes in object state (e.g., color, position) within a scene. \\ \cline{2-5} 
    
    & \textbf{Depth \& Distance} & Playing4Data / POEM-V2 & 300 & Evaluates the model's ability to understand depth perception and distance relationships between objects in a given environment. \\ \cline{2-5} 

    & \textbf{Action Prediction} & WM-ABench & 100 & Assesses the model’s ability to predict the next action or determine task completion based on current spatial descriptions. \\ \hline
    
    \multirow{4}{=}{\centering \textbf{Logic Detection}} 
    & \textbf{Object Peranence} & CoSpace & 200 & Tests the model's understanding of object permanence, evaluating its ability to reason about objects' existence across different views. \\ \cline{2-5} 
    
    & \textbf{Direction Judgement} & CoSpace & 250 & Assesses the model's ability to determine cardinal directions (e.g., east, west) based on spatial information in images or descriptions. \\ \cline{2-5} 
    
    \hline
    \end{tabular}
    } 
    
    \caption{\textbf{The task classification, task description and data sources of SiT-bench.}}
    \label{tab:category}
\end{table*}

\twocolumn\subsection{DeepSeek-R1 Filter Prompt Template}
\label{app:filtering_protocol}

To ensure the quality of our benchmark, we employ DeepSeek-R1 as an automated data quality auditor. The filtering process is guided by five core principles:

\begin{enumerate}[leftmargin=*]
    \item \textbf{Entity Visibility \& Presence:} If the captions across all images fail to explicitly mention the primary entities or objects essential for answering the question, the data item is deemed unusable.
    
    \item \textbf{No Answer Leakage:} If the question itself nearly contains or directly reveals the answer, requiring no spatial inference to complete the task, the data item is discarded.
    
    \item \textbf{Spatial Deductibility:} We ensure that the directional terms provided in captions (e.g., ``left of'', ``behind'') are sufficient to logically determine a unique answer. Items with overly vague descriptions are removed.
    
    \item \textbf{Multi-View Reasoning Priority:} We prioritize items that require integrating information from multiple images to solve, as this represents high-quality 3D spatial reasoning.
    
    \item \textbf{Ambiguity Detection:} Items that may yield multiple reasonable interpretations based on the textual description are excluded.
\end{enumerate}

The complete prompt template used for DeepSeek-R1 filtering is presented below:

\begin{tcolorbox}[prompt={DeepSeek-R1 Filtering Prompt},breakable]
Act as a strict data quality auditor for a 3D spatial reasoning dataset. You are provided with:\\
1. \textbf{Captions}: Detailed descriptions of multiple images from the same scene.\\
2. \textbf{Question}: A spatial reasoning task (e.g., relative direction) based on those images.\\
3. \textbf{Choices}: Multiple-choice options.\\
4. \textbf{Answer}: The ground truth answer.\\
\newline
Your goal is to identify high-quality spatial reasoning items. A high-quality item must satisfy ALL following criteria:\\
\newline
\textbf{1. Entity Visibility \& Presence (STRICT):}\\
- All objects mentioned in the question (the `standing at' object, the `facing' object, and the `target' object) MUST be explicitly described in at least one of the image captions.\\
- If a caption mentions an object is ``not visible'' or ``not found'', and no other image provides its description, the item MUST be DISCARDED.\\
\newline
\textbf{2. No Answer Leakage \& Triviality:}\\
- The question or choices must NOT make the answer obvious without the captions.\\
- The captions must NOT explicitly state the answer (e.g., if the question is ``where is the lamp relative to the desk'', a caption saying ``the lamp is on the right of the desk'' is a direct leak).\\
- The task should not be solvable by common sense alone (e.g., ``Where is the ceiling? A. Up'').\\
\newline
\textbf{3. Spatial Deductibility (Multi-View Reasoning Priority):}\\
- The answer must be logically derivable from the provided spatial relationships (left, right, behind, front, adjacent, etc.).\\
- Prefer items that require combining information from multiple image captions to solve the 3D layout.\\
- If the captions are too vague or contradictory, DISCARD the item.\\
\newline
\textbf{4. Ambiguity Check:}\\
- The correct choice should be the only logically sound answer based on the text.\\
\newline
Output your decision in EXACTLY this JSON format:\\
\{\\
\hspace*{1em}``keep'': true or false,\\
\hspace*{1em}``reason'': ``Explain your logic: 1) Are all entities present? 2) Is there leakage? 3) Is it logically solvable?'',\\
\hspace*{1em}``missing\_entities'': [``list any entities from the question not found in captions''],\\
\hspace*{1em}``is\_multi\_view'': true or false,\\
\hspace*{1em}``leakage\_detected'': true or false\\
\}
\end{tcolorbox}

\subsection{Detailed Implementation Parameters}
\label{app:implement_details}

All experiments are conducted on a server equipped with 8$\times$NVIDIA A100 GPUs (80GB each). For models with fewer than 100B parameters, we employ vLLM (version 0.11.0) as the inference backend to enable efficient batched generation. The detailed configuration parameters are summarized in Table~\ref{tab:impl_params}.

\begin{table}[h]
\centering
\small
\begin{tabular}{l l}
\toprule
\textbf{Parameter} & \textbf{Value} \\
\midrule
\multicolumn{2}{l}{\textit{Server Configuration}} \\
Tensor Parallelism (TP) & Model-dependent \\
GPU Memory Utilization & Auto \\
Max Batch Size & 32 \\
Max Model Length & Model-dependent \\
KV Cache Dtype & Auto \\
\midrule
\multicolumn{2}{l}{\textit{Generation Parameters}} \\
Temperature & 0.0 \\
Top-p & 1.0 \\
Max Tokens & 32,768 \\
Presence Penalty & 0.0 \\
Frequency Penalty & 0.0 \\
Repetition Penalty & 1.1 \\
\midrule
\multicolumn{2}{l}{\textit{Inference Settings}} \\
Concurrency & 16 \\
Timeout (seconds) & 1,200 \\
Thinking Budget & 8,192 \\
Answer Format & Plain \\
\bottomrule
\end{tabular}
\caption{Implementation parameters for SiT-Bench evaluation.}
\label{tab:impl_params}
\end{table}

For reasoning-enhanced evaluation (CoT prompting), we enable the thinking mode with a budget of 8,192 tokens. All API-based models (e.g., GPT-4o, Gemini-3-Flash) are accessed through their official endpoints with default rate limits. For open-source models, we deploy local vLLM servers with OpenAI-compatible APIs, using trust remote code when necessary for custom architectures.

\subsection{Complete Gemini-3-Flash Reasoning Process}
\label{app:reasoning_process}

In this section, we present the complete reasoning process of Gemini-3-Flash on two representative examples from our benchmark. These examples demonstrate the model's step-by-step spatial reasoning capabilities.

\begin{tcolorbox}[prompt={Gemini-3-Flash Reasoning Process - Example 1},breakable]
\textbf{Extended Thinking Process:}\\[0.5em]
\textit{Visualizing the Layout}\\
I'm currently trying to establish the spatial relationships between the provided images. The initial focus is on correctly identifying the cardinal directions. I've placed Image 1, featuring the silver sedan and red cement mixer, as representing North, as a starting point. Then, I'll assess the remaining three images and will position them according to their orientation to this central image.\\[0.3em]

\textit{Orienting the Images}\\
I've mapped out the cardinal directions and tentatively placed the images. Image 1 represents North, as initially hypothesized. Now, I'm analyzing the relationships of vehicles across images. The silver sedan appears in multiple locations, which makes me think of possible multiple instances or, more likely, a shared space with a rotating viewpoint.\\[0.3em]

\textit{Mapping Vehicle Positions}\\
I'm now cross-referencing vehicle locations across the four images to solidify their relative positions. It seems certain vehicles are present in multiple images, such as the silver sedan and the red cement mixer. Analyzing the mixer's changing position across North, East, and West supports the initial orientation. Additionally, I am trying to determine if all mentions are the same vehicle. I am making progress.\\[0.3em]

\textit{Confirming Vehicle Locations}\\
I'm now carefully tracking the positions of each vehicle across the four images to solidify the layout. Analyzing the silver sedan is tricky, as it appears in three of the four images, but its position shifts. I am now certain the silver sedan is NOT a single instance, and I need to note this variance. The red cement mixer seems to be a stable element.\\[0.3em]

\textit{Revisiting the Vehicles}\\
I'm now carefully re-examining the vehicles, specifically the silver sedan. Initially, I thought it was a single instance, but its shifting positions in Images 1, 2, and 3 suggest otherwise. I'm focusing on the silver sedan's location relative to the red cement mixer to define the spatial relationships. This will help with the layout's finalization.\\[0.3em]

\textit{Refining Vehicle Identification}\\
I'm now certain there are two distinct silver sedans. Image 1's sedan is different from the one in Images 2 and 3. My focus is now the vehicles' positions relative to the cardinal directions, and how their positions change across the images to establish a more concrete layout.\\[0.3em]

\textit{Mapping the Scene}\\
I've finally pinpointed the position of both silver sedans, and the red mixer, which aids greatly in establishing the layout. I've re-mapped the vehicles to reflect the cardinal directions of each image, which solidifies the spatial relationships between the vehicles. It is clear that the silver sedans occupy opposing sides of the scene.\\[0.3em]

\textit{Finalizing Vehicle Locations}\\
I've carefully separated the two silver sedans and noted their positions in the four images. One is consistently located on the west side of the scene, while the other is on the east. The red cement mixer serves as a central point, further solidifying the layout and confirming the relationships between the cardinal directions.\\[0.3em]

\textit{Analyzing the Vehicles}\\
I'm now certain there are two distinct silver sedans. My current focus is refining the spatial layout by cross-referencing the red cement mixer and both silver sedans to map their location in each image, and noting the SUV. This will help with the final spatial determination.\\[0.3em]

\textit{Re-evaluating the Layout}\\
I'm now certain there are two distinct silver sedans. The silver sedan in Image 1, on the west, is likely parked near the building. The second silver sedan, found in Images 2 and 3, is located further east. I'm noting the changes in the mixer and SUV in addition to the sedan's locations.\\[0.3em]

\textit{Confirming the Silver Sedans}\\
I'm now sure there are two separate silver sedans, and their positions relative to each other and the mixer clarify the scene layout. It is clear that one sedan is near a building to the west, while the other is in the east. The red cement mixer forms a central point.\\[0.3em]

\textit{Mapping the Scene}\\
I've now cross-referenced the locations of the red cement mixer, the black/dark SUV, and the two silver sedans. The red mixer's placement aids in establishing the cardinal directions, as do the SUV's and the sedans' positions across the images.\\[0.3em]

\textit{Mapping the Scene}\\
I've re-examined the vehicles and the cardinal directions to establish a spatial framework. With the two distinct silver sedans now placed, I'm using the cement mixer as the key for orienting the scene across images. I'm noting the changes in the SUV's position as well.\\[0.3em]

\textit{Refining Vehicle Locations}\\
I'm now certain there are two distinct silver sedans. Based on the position of the red mixer and the SUV, I've confirmed their general placement, allowing me to refine the positions of both silver sedans relative to the cardinal directions. The goal is to accurately map the vehicles' locations across each image.\\[0.3em]

\textit{Mapping the Scene}\\
I've re-examined the vehicles and the cardinal directions to form a spatial framework. Having distinguished the two silver sedans, and the red mixer, I'm using the cement mixer and SUV as keys to orient the scene across images. I'm noting the changes in the SUV's position as well.\\[0.3em]

\textit{Reassessing the Scene}\\
I've re-examined the vehicles and their positions. I'm now certain of two distinct silver sedans and their locations. One's West, near a building; the other's East, with the red cement mixer serving as a key reference point. The SUV's location further confirms the layout.\\[0.3em]

\textit{Mapping the Scene}\\
I've re-examined the vehicles and the cardinal directions. One silver sedan is in the west, likely near a building. Another is in the east. The red cement mixer and the dark SUV help orient the scene.\\[0.3em]

\textit{Analyzing Vehicle Locations}\\
I'm now carefully tracking the positions of each vehicle in each image. Both silver sedans are mapped out. I've re-examined the SUV's and the cement mixer's relative positions, with an eye toward solidifying the layout and ensuring I have identified all the unique vehicles.\\[0.3em]

\textit{Revising Cardinal Directions}\\
I'm now certain the directions are as I've laid them out. My immediate concern is the relationship between the two silver sedans and other vehicles. I'll need to re-map the scene and incorporate the SUV to verify everything.\\[0.3em]

\textit{Reassessing the Layout}\\
I am now thoroughly examining the placement of each vehicle across the four images to solidify the layout. After verifying that Image 1 faces North, I'm focusing on the SUV and the red cement mixer for orientation. The key is to correctly place the two silver sedans.\\[0.3em]

\textit{Mapping the Scene}\\
I am now carefully using the descriptions to map the placement of all the vehicles. My goal is now to solidify the layout based on vehicle locations across each image, including the two silver sedans and red cement mixer. Analyzing the SUV and the red brick building will help greatly.\\[0.3em]

\textit{Refining the Layout}\\
I'm now zeroing in on the spatial arrangement, focused on the relationships between the two silver sedans, the red mixer, and the SUV. The varying perspectives have made this challenging. I believe I have the primary positions. I am making rapid progress in confirming this.\\[0.3em]

\textit{Analyzing Vehicle Placement}\\
I'm now certain of the two silver sedans and their locations. One is in the west, likely near a building. The other is consistently east. I'm focusing on the relationship between these sedans, the red cement mixer, and the dark SUV to refine the image's orientation and solidify the spatial layout.\\[0.3em]

\textit{Mapping the Scene}\\
I am now carefully using the descriptions to map the placement of all the vehicles. My goal is now to solidify the layout based on vehicle locations across each image, including the two silver sedans and red cement mixer. Analyzing the SUV and the red brick building will help greatly.\\[0.3em]

\textit{Refining the Layout}\\
I'm now certain there are two distinct silver sedans. Based on the position of the red mixer and the SUV, I've refined the vehicle placements. Image 1 features one sedan to the West, and Image 2 shows the other sedan to the East. I'm building out a map.\\[0.3em]

\textit{Re-examining Vehicle Locations}\\
I'm making a final push to solidify the vehicle locations. Image 1 faces North, and the red cement mixer is central, supporting this initial orientation. The varying positions of both silver sedans, and the shared appearance of the red cement mixer in multiple images, help confirm the spatial layout.\\[0.3em]

\textit{Refining the Layout}\\
I'm now certain the directions are as I've laid them out, and I can start solidifying the layout. My current focus is how the silver sedans and other vehicles relate. The SUV's position is a key element and aids my understanding of the cardinal directions of each image.\\[1em]

\textbf{Final Reasoning (Think Block):}\\[0.5em]
The user wants to know the total number of vehicles in the scene based on four image captions.

1. Analyze Image 1 (North): Silver sedan (far left - West), Red cement mixer (center - North), Black SUV (far right - East). Vehicles identified: Sedan A (West), Mixer (North), SUV (East).

2. Analyze Image 2 (Direction to be deduced): Red cement mixer (far left), Silver sedan (center), Dark-colored SUV (right). Deduction: If the Mixer is North and it's on the left, and the SUV is East and it's on the right, this image is likely facing East. Vehicles identified: Mixer (North), Sedan B (East), SUV (East/South-East).

3. Analyze Image 3 (Direction to be deduced): Silver sedan (far left), Red brick building (center), Large tree (far right). Deduction: If the Silver Sedan B is in the East, and it's on the far left, this image is likely facing South. Vehicles identified: Sedan B (East).

4. Analyze Image 4 (Direction to be deduced): Black metal fence (far left), Red brick building under construction (center), Red cement mixer (far right). Deduction: If the Mixer is North and it's on the far right, this image is likely facing West. Vehicles identified: Mixer (North).

5. Consolidate the vehicles:
   - Vehicle 1: Silver sedan (West) - from Image 1 (far left).
   - Vehicle 2: Red cement mixer (North) - from Image 1 (center), Image 2 (far left), Image 4 (far right).
   - Vehicle 3: Black/Dark SUV (East/South-East) - from Image 1 (far right), Image 2 (right).
   - Vehicle 4: Silver sedan (East) - from Image 2 (center), Image 3 (far left).

Total distinct vehicles: 4.\\[0.5em]

\textbf{Answer:} 4
\end{tcolorbox}

\begin{tcolorbox}[prompt={Gemini-3-Flash Reasoning Process - Example 2},breakable]
\textbf{Question:} Based on these four images (image 1, 2, 3, and 4) showing the scene from different viewpoints (north, west, south, and east), with each camera aligned with room walls and partially capturing the surroundings: How many motorcycles are there in total in the scene?\\[1em]

\textbf{Extended Thinking Process:}\\[0.5em]

\textit{Analyzing the Visual Data}\\
I'm currently focused on the images and their composition. Image 1 is set, showing a black motorcycle, a yellow bicycle, and a leafless tree on the far left. To the right, another dark motorcycle is partially obscured. Further along, a gray building with AC units and a third motorcycle near a yellow sign is visible. I'm noting the positions of the objects and their relative distance from each other.\\[0.3em]

\textit{Assessing Image 2's Contents}\\
I'm now carefully examining Image 2, trying to catalog the scene. I see a silver sedan, then a black motorcycle with a rear rack, and a second one near a leafless tree. There are also bollards, parked cars, and a brown brick wall. The ``dark two-wheeler'' warrants closer scrutiny, it appears I was correct - it's another motorcycle.\\[0.3em]

\textit{Mapping the Image Directions}\\
The images are now aligned. Image 1 is north, with a good view. I've broken down each image, itemizing motorcycles. Image 2 also provides insight with its silver sedan, and the positioning of a second black motorcycle. Image 3 seems to include one motorcycle and Image 4 appears motorcycle-free.\\[0.3em]

\textit{Deducing Image Directions}\\
I've made progress in establishing the directions. Confirmed that Image 2 looks West, relative to Image 1 (North). Analyzing the objects in Images 1 and 2, and their relative positions, has clarified the directional relationships. I'm now proceeding to determine the orientation of Images 3 and 4, and the relative positions of the buildings.\\[0.3em]

\textit{Revisiting Directional Overlaps}\\
I've just refined my focus, cross-referencing objects across images to pinpoint directions. The presence of the yellow bike and leafless tree has definitively established that Image 2 is looking West. I'm now comparing the positioning of the gray building and motorcycles to confirm the orientation of Images 3 and 4, building up a cohesive map.\\[0.3em]

\textit{Refining the Overlaps}\\
I've clarified the image directions by cross-referencing features. Image 2's West direction is confirmed due to the matching of motorcycles and the yellow bike/tree, consistent with Image 1's North view. Now, I'm using the gray building and stone stairs to definitively establish that Image 4 is East, followed by Image 3's placement.\\[0.3em]

\textit{Confirming Image Orientations}\\
I'm now verifying the orientations based on object overlaps. The tree and bike confirm Image 2 as West. The stone stairs in both Image 3 and Image 4 suggest that Image 4 is East. I'm now certain that the remaining direction is South, and verifying motorcycle counts.\\[0.3em]

\textit{Determining the Coordinates}\\
I'm now carefully mapping the image orientations. After confirming the West direction with the tree/bike overlap, I identified East using the gray building and stone stairs. This means that Image 3 is south, and I'm now proceeding to reconcile the motorcycle counts across images, and pinpointing their exact locations.\\[0.3em]

\textit{Reconciling Motorcycle Counts}\\
I'm now fully immersed in reconciling motorcycle counts across the images. I've re-listed all the bikes, accounting for overlaps based on their positions. I've pinpointed their locations based on the image orientations, identifying the same bikes in multiple views. I'm building a concise summary of motorcycle positions.\\[0.3em]

\textit{Summarizing the Findings}\\
I'm now fully immersed in the motorcycle analysis. Based on object overlaps and directional clarity, I've confirmed that the `dark two-wheeler' and the first black bike are the same motorcycle. I've streamlined my focus on the motorcycle counts, their locations based on the image orientations, and the building layouts.\\[0.3em]

\textit{Summarizing the Motorcycle Positions}\\
I'm now consolidating the motorcycle locations based on the analysis. The black bike with a rear rack is likely the same motorcycle seen on Image 3 (with the handlebars visible). I'm also confident in the placement of the other motorcycles. I am finishing the final summary of motorcycle counts.\\[0.3em]

\textit{Confirming the Final Counts}\\
I'm now concluding the analysis, confident in the final motorcycle counts. The black motorcycle with the rear rack and the motorcycle with handlebars are the same. Image 1 has 3, Image 2 has 3, Image 3 has 1, and Image 4 has 0. I will now finalize my overall assessment.\\[0.3em]

\textit{Re-evaluating Motorcycle Counts}\\
I'm now revising the motorcycle counts, re-examining the images. I've re-identified the motorcycles in Image 1: 3. Then in Image 2: I am seeing 3. Image 3 has one. Image 4 has zero. Next step, building my overall findings.\\[0.3em]

\textit{Refining the Motorcycle Data}\\
I'm now consolidating the motorcycle locations based on the analysis. The black bike with a rear rack is likely the same motorcycle seen on Image 3 (with the handlebars visible). The yellow bike/tree cross-reference confirms the overall orientation. I am finishing the final summary of motorcycle counts and positions.\\[0.3em]

\textit{Concluding the Motorcycle Counts}\\
I'm now concluding the analysis, confident in the final motorcycle counts. The black motorcycle with the rear rack and the motorcycle with handlebars are the same. Image 1 has 3, Image 2 has 3, Image 3 has 1, and Image 4 has 0. I will now finalize my overall assessment.\\[0.3em]

\textit{Re-examining Motorcycle Counts}\\
I'm revisiting the initial motorcycle count, cross-referencing descriptions. I've re-examined the images to ensure consistency. Image 1 has three motorcycles, with a ``dark two-wheeler'' in Image 2. I'm building a solid foundation of data.\\[1em]

\textbf{Final Reasoning (Think Block):}\\[0.5em]
The user wants to know the total number of motorcycles in the scene based on four image captions.

1. Analyze Image 1 (North): Motorcycle 1 (M1): Black, far left, near metal railing, yellow bicycle, and leafless tree. Motorcycle 2 (M2): Dark-colored, behind the tree. Motorcycle 3 (M3): Near yellow-and-black traffic cones, further right. Total in Image 1: 3 motorcycles.

2. Analyze Image 2 (West): Motorcycle 4 (M4): Black, rear rack, beside a dark sedan, near a silver sedan and bollards (left side). Motorcycle 2 (M2): Black, next to a leafless tree (right side). Motorcycle 1 (M1): ``Another dark two-wheeler'' beside a yellow bicycle and the tree (right side). Total in Image 2: 3 motorcycles (M4, M2, M1).

3. Analyze Image 3 (South): Motorcycle 4 (M4): Black, handlebars/front wheel visible, next to a black bicycle and bollards (right side). Total in Image 3: 1 motorcycle (M4).

4. Analyze Image 4 (East): No motorcycles mentioned.

5. Consolidate and Count: North-West Corner: M1 and M2 are mentioned in both Image 1 (left) and Image 2 (right). North-East Corner: M3 is mentioned in Image 1 (right). South-East Corner: No motorcycles mentioned.

Total unique motorcycles: M1, M2, M3.
Total count = 3.\\[0.5em]

\textbf{Answer:} 3
\end{tcolorbox}

\subsection{Instructions to Human Test Participants}
\label{app:human_instructions}

In this section, we present the complete instructions provided to human participants during our benchmark evaluation. These instructions were designed to ensure consistent and fair comparison between human and model performance.

\begin{tcolorbox}[prompt={Instructions for Human Participants},breakable]
\textbf{Welcome to the Spatial Intelligence Test}\\[0.5em]

Thank you for participating in this evaluation. You will be presented with a series of questions that assess your spatial reasoning abilities based on visual information.\\[0.5em]

\textbf{General Guidelines:}\\[0.3em]
\begin{enumerate}
    \item \textbf{Read Carefully:} Each question will present multiple images showing different viewpoints of a scene or object. Read the question thoroughly before examining the images.
    
    \item \textbf{Examine All Images:} Take time to study each provided image. Pay attention to spatial relationships, object positions, and directional cues.
    
    \item \textbf{No External Tools:} Do not use any external tools, calculators, or reference materials. Rely solely on your visual perception and spatial reasoning.
    
    \item \textbf{Time Limit:} There is no strict time limit per question, but please work at a steady pace. The average completion time is approximately 45-60 minutes for the full test.
    
    \item \textbf{Single Attempt:} Select the answer you believe is most correct. You cannot change your answer after submission.
    
    \item \textbf{Best Effort:} If you are uncertain, make your best educated guess rather than leaving the question blank.
\end{enumerate}

\textbf{Question Types:}\\[0.3em]
You will encounter various types of spatial reasoning questions, including:
\begin{itemize}
    \item \textbf{Scene Layout:} Determining spatial relationships between objects from multiple viewpoints.
    \item \textbf{Object Counting:} Identifying and counting specific objects across different views.
    \item \textbf{Direction \& Orientation:} Understanding cardinal directions and relative positions.
    \item \textbf{Spatial Transformation:} Mental rotation and perspective-taking tasks.
\end{itemize}

\textbf{Important Notes:}\\[0.3em]
\begin{itemize}
    \item All images are captured from real-world scenes or carefully constructed environments.
    \item Cardinal directions (North, South, East, West) may be indicated in some questions.
    \item When multiple images are provided, they typically show the same scene from different angles.
    \item Pay attention to overlapping objects between images to establish spatial consistency.
\end{itemize}

\textbf{Example Approach:}\\[0.3em]
When solving a multi-view spatial question:
\begin{enumerate}
    \item Identify a reference object visible in multiple images.
    \item Establish the viewing direction for each image.
    \item Build a mental map of the scene layout.
    \item Use this mental map to answer the question.
\end{enumerate}

\textbf{Ready to Begin?}\\[0.3em]
Please ensure you are in a quiet environment with minimal distractions. Click ``Start'' when you are ready to proceed.\\[0.5em]

\textit{Thank you for your participation. Your responses will contribute to important research in spatial intelligence evaluation.}
\end{tcolorbox}

\subsection{More Models' Complete Evaluation Results on SiT-Bench}
\label{app:more_results}

This section presents comprehensive evaluation results for additional models on the SiT-Bench benchmark. The complete performance metrics across all spatial reasoning tasks are provided in table~\ref{tab:more_results}.

\renewcommand{\arraystretch}{1.25} 
\setlength{\tabcolsep}{3pt} 
\definecolor{lightyellow}{RGB}{255, 250, 194}
\definecolor{lightred}{RGB}{250, 189, 164}
\definecolor{lightorange}{RGB}{254, 226, 205}
\begin{table*}[ht]
\footnotesize
    \centering
    \small
    \resizebox{1.0\textwidth}{!}{
    \begin{tabular}{l| c c| c c c c| c c c c| c c c c c c| c c c c c| c c c}
        \textbf{Models} & \rotatebox{75}{\textbf{Rank}} & \normalsize\rotatebox{75}{\textbf{Avg.}} & \normalsize\rotatebox{75}{\makecell{\textbf{Global Perception} \\ \textbf{\& Mapping}}} & 
       \scriptsize \rotatebox{75}{Scene Layout Reason} & \scriptsize \rotatebox{75}{Panoramic Counting} & \scriptsize \rotatebox{75}{Cognitive Mapping} & \normalsize\rotatebox{75}{\makecell{\textbf{Navigation} \\ \textbf{\& Planning}}} & \scriptsize \rotatebox{75}{Outdoor Navigation} & \scriptsize \rotatebox{75}{Path Planning Logic} & \scriptsize \rotatebox{75}{Ego/Objects-motion Perception} & \normalsize\rotatebox{75}{\makecell{\textbf{Multi-View \&} \\ \textbf{Geometric Reasoning}}} & \scriptsize \rotatebox{75}{Real-world QA} & \scriptsize \rotatebox{75}{View Consistency} & \scriptsize \rotatebox{75}{Perspective Shift} & \scriptsize \rotatebox{75}{Pure Mental Rotation} & \scriptsize \rotatebox{75}{Spatial Puzzles}& \normalsize\rotatebox{75}{\makecell{\textbf{Embodied \&} \\ \textbf{Fine-grained}}} & \scriptsize \rotatebox{75}{Hand-Object Interaction} & \scriptsize \rotatebox{75}{Fine-grained Tracking} & \scriptsize\rotatebox{75}{Depth \& Distance} & \scriptsize \rotatebox{75}{Action Prediction} & \normalsize\rotatebox{75}{\textbf{Logic Detection}} & \scriptsize \rotatebox{75}{Object Peranence} & \scriptsize\rotatebox{75}{Direction Judgement}\\
        \hline
        \textbf{Baseline} & & & & & & & & & & & & & & & & & & & & & & & & \\
        Human Level & 1 & 74.42 & 67.85 & 80.00 & 73.42 & 26.77 & 78.22 & 64.67 & 95.00 & 83.00 & 77.45 & 98.51 & 75.00 & 71.23 & 68.00 & 93.00 & 71.86 & 71.50 & 72.13 & 77.67 & 55.00 & 76.22 & 70.00 & 81.20 \\
        Random Level & 32 & 27.30 & - & 25.00 & 25.00 & - & 34.72 & 12.50 & 25.00 & 50.00 & 24.99 & 24.96 & 25.00 & 25.00 & 25.00 & 25.00 & 24.98 & 24.95 & 25.00 & 25.00 & 25.00 & 25.00 & 25.00 & 25.00 \\
        \hline
        \textbf{Proprietary Models / 100B+ Models} & & & & & & & & & & & & & & & & & & & & & & & & \\
        GPT-4o~\cite{hurst2024gpt} & 6 & \cellcolor{lightyellow} 45.70 & 17.74 & 11.50 & 26.58 & 3.61 & 53.78 & 32.00 & 85.00 & 60.60 & \cellcolor{lightyellow} 54.55 & 91.85 & 39.00 & 51.28 & 37.00 & 74.00 & \cellcolor{lightyellow} 47.78 & 30.00 & 56.07 & 70.67 & 25.00 & 45.33 & 74.00 & 22.40 \\
        DeepSeek-V3.2~\cite{liu2025deepseek} & 22 & 37.06 & 19.68 & 13.50 & 29.24 & 3.30 & 49.89 & 19.67 & 87.00 & 60.60 & 46.65 & 93.33 & 38.00 & 33.05 & 36.00 & 72.00 & 33.67 & 29.50 & 39.34 & 38.00 & 20.00 & 25.11 & 21.50 & 28.00 \\
        ~~~-thinking & 10 & 43.74 & \cellcolor{lightyellow} 22.02 & 16.50 & 32.89 & 0.33 & \cellcolor{lightyellow} 61.22 & 12.00 & 86.00 & 85.80 & 53.71 & 97.78 & 37.00 & 47.29 & 37.00 & 80.00 & 32.76 & 13.25 & 55.08 & 37.33 & 29.00 & \cellcolor{lightyellow} 46.22 & 63.00 & 32.80\\
        Gemini-3-Flash-preview~\cite{gemini3flash} & 2 & \cellcolor{lightred} 59.46 & \cellcolor{lightred} 35.66 & 44.50 & 38.87 & 8.34 & \cellcolor{lightred} 77.11 & 47.00 & 89.00 & 92.80 & \cellcolor{lightred} 68.54 & 96.30 & 50.50 & 72.65 & 45.00 & 84.00 & \cellcolor{lightred} 51.31 & 27.75 & 65.25 & 76.67 & 27.00 & \cellcolor{lightred} 59.11 & 61.00 & 57.60 \\
        \hline
        \textbf{Open-Source Models / 100B- Models} & & & & & & & & & & & & & & & & & & & & & & & & \\
        LlaVA-1.5-7B~\cite{liu2024improved} & 31 & 30.53 & \cellcolor{lightred} 29.18 & 28.00 & 39.53 & 0.34 & 39.33 & 16.33 & 95.00 & 42.00 & 29.78 & 28.89 & 31.00 & 31.91 & 25.00 & 22.00 & 25.52 & 23.25 & 30.16 & 22.33 & 30.00 & 28.44 & 22.50 & 33.20 \\
        Llama-3.1-8B~\cite{grattafiori2024llama} & 27 & 34.78 & 14.28 & 15.00 & 17.94 & 1.82 & 45.11 & 17.00 & 71.00 & 56.80 & 36.60 & 88.15 & 31.50 & 21.94 & 27.00 & 40.00 & 39.73 & 51.75 & 34.43 & 31.33 & 33.00 & 26.00 & 19.50 & 31.20 \\
        InternVL3-2B~\cite{zhu2025internvl3} & 29 & 33.92 & 20.68 & 16.50 & 29.90 & 1.28 & 42.67 & 18.00 & 87.00 & 48.60 & 39.59 & 87.41 & 32.00 & 30.48 & 24.00 & 36.00 & 31.22 & 6.75 & 36.39 & 24.67 & 13.00 & 30.22 & 22.50 & 36.40 \\
        InternVL3-8B~\cite{zhu2025internvl3} & 20 & 38.42 & 22.68 & 13.50 & 35.88 & 1.29 & 35.00 & 10.33 & 71.00 & 42.60 & 46.41 & 93.33 & 33.50 & 38.18 & 27.00 & 68.00 & 47.06 & 53.75 & 46.56 & 43.33 & 33.00 & 30.22 & 29.00 & 31.20 \\
        InternVL3-14B & - & 42.58 & 20.19 & 11 & 31.89 & 3.32 & 48.89 & 20.67 & 89 & 57.8 & 45.69 & 97.04 & 35 & 33.05 & 30 & 70 & 49.59 & 45.5 & 51.15 & 59.33 & 32 & 36.89 & 41 & 33.6 \\
        InternVL3\_5-2B & - & 34.75 & 24.19 & 11 & 39.87 & 3.38 & 45.56 & 13.67 & 88 & 56.2 & 41.87 & 87.41 & 39 & 32.19 & 23 & 36 & 30.68 & 32 & 35.74 & 27.67 & 19 & 24 & 17.5 & 29.2 \\
        InternVL3.5-4B~\cite{wang2025internvl3} & 17 & 39.95 & 25.79 & 15.00 & 40.20 & 4.01 & 47.44 & 17.67 & 88.00 & 57.20 & 44.50 & 95.56 & 35.50 & 32.48 & 28.00 & 60.00 & 38.73 & 36.50 & 43.93 & 38.00 & 34.00 & 38.44 & 46.50 & 32.00 \\
        ~~~-thinking & 18 & 38.98 & 22.14 & 17.50 & 32.23 & 1.04 & 47.00 & 21.33 & 77.00 & 56.40 & 40.43 & 93.33 & 30.00 & 27.92 & 23.00 & 62.00 & 38.10 & 36.25 & 48.52 & 30.33 & 37.00 & 44.89 & 57.00 & 35.20 \\
        InternVL3.5-8B~\cite{wang2025internvl3} & 12 & 43.27 & 26.14 & 18.50 & 38.87 & 3.09 & 49.78 & 19.33 & 90.00 & 60.00 & 44.26 & 94.81 & 34.50 & 33.05 & 26.00 & 62.00 & 48.78 & 48.00 & 52.46 & 49.33 & 39.00 & 37.78 & 45.50 & 31.60 \\
        ~~~-thinking & 4 & \cellcolor{lightyellow} 46.43 & 18.65 & 14.50 & 27.24 & 1.07 & \cellcolor{lightyellow} 62.00 & 24.33 & 85.00 & 80.00 & 52.87 & 96.30 & 39.00 & 46.72 & 28.00 & 84.00 & 45.61 & 42.75 & 57.05 & 44.00 & 27.00 & 42.44 & 52.50 & 34.40 \\
        InternVL3\_5-14B-Instruct & - & 41.47 & 23.64 & 16.5 & 35.55 & 2.05 & 49.33 & 23.67 & 88 & 57 & 46.17 & 95.56 & 35 & 34.47 & 34 & 64 & 42.81 & 34 & 47.54 & 56 & 24 & 37.56 & 44 & 32.4 \\
        InternVL3\_5-30B-A3B & - & 40.98 & 20.32 & 12.5 & 30.23 & 6.12 & 47.44 & 19 & 88 & 56.4 & 53.59 & 97.04 & 36.5 & 50.43 & 31 & 72 & 40.54 & 33.5 & 41.31 & 49 & 41 & 33.33 & 30 & 36 \\
        
        Qwen2.5-3B~\cite{yang2024qwen2} & 26 & 34.81 & \cellcolor{lightyellow} 27.93 & 19.50 & 42.52 & 0.83 & 45.44 & 15.67 & 75.00 & 57.40 & 35.05 & 83.70 & 32.50 & 19.66 & 32.00 & 28.00 & 32.85 & 29.25 & 38.03 & 36.00 & 22.00 & 27.11 & 18.00 & 34.40 \\
        Qwen2.5-7B & - & 36.43 & 12.77 & 16 & 14.62 & 0.77 & 41.22 & 15.67 & 84 & 48 & 45.81 & 85.93 & 32 & 46.44 & 22 & 36 & 40.36 & 43.75 & 40.66 & 38.67 & 31 & 31.33 & 27.5 & 34.4 \\
        Qwen2.5-72B~\cite{yang2024qwen2} & 13 & 42.57 & 14.28 & 15.00 & 17.94 & 1.84 & 50.56 & 26.33 & 90.00 & 57.20 & 53.23 & 95.56 & 39.50 & 48.43 & 35.00 & 64.00 & 47.15 & 36.75 & 43.61 & 70.00 & 31.00 & 33.33 & 32.00 & 34.40 \\
        Qwen2.5-VL-3B~\cite{bai2025qwen2} & 25 & 35.54 & 21.49 & 10.00 & 35.22 & 3.17 & 40.89 & 11.33 & 79.00 & 51.00 & 40.55 & 91.85 & 36.50 & 27.07 & 27.00 & 40.00 & 39.10 & 48.00 & 33.44 & 34.00 & 36.00 & 25.56 & 18.00 & 31.60 \\
        Qwen2.5-VL-7B & - & 34.7 & 19.9 & 15.5 & 27.57 & 5.61 & 42.78 & 15 & 62 & 55.6 & 46.29 & 94.81 & 35 & 37.32 & 29 & 58 & 32.67 & 23.5 & 43.61 & 37.33 & 22 & 21.78 & 28.5 & 16.4 \\
        Qwen2.5-VL-72B~\cite{bai2025qwen2} & 8 & 45.45 & 19.29 & 13.00 & 28.90 & 2.94 & 55.67 & 33.33 & 89.00 & 62.40 & \cellcolor{lightyellow} 53.47 & 95.56 & 36.50 & 48.43 & 38.00 & 74.00 & \cellcolor{lightyellow} 49.59 & 33.00 & 52.79 & 76.00 & 27.00 & 34.89 & 47.50 & 24.80 \\
        Qwen3-4B-Instruct-2507 & - & 36.59 & 18.95 & 14 & 27.91 & 1.91 & 44.22 & 16.67 & 69 & 55.8 & 40.91 & 89.63 & 34 & 28.77 & 23 & 58 & 38.1 & 47.75 & 34.43 & 32 & 29 & 33.11 & 30.5 & 35.2 \\
        Qwen3-30B-A3B-Instruct-2507 & - & 36.5 & 18.38 & 16.5 & 24.92 & 2.49 & 42 & 15.67 & 92 & 47.8 & 46.41 & 92.59 & 37 & 35.9 & 32 & 62 & 37.38 & 33.75 & 36.39 & 41 & 44 & 29.11 & 20.5 & 36 \\
        Qwen3-4B~\cite{yang2024qwen2} & 28 & 34.68 & 12.44 & 16.00 & 13.29 & 2.76 &  45.89 & 20.33 & 84.00 & 53.60 & 43.06 & 87.41 & 37.50 & 34.76 & 25.00 & 40.00 & 34.57 & 38.25 & 36.07 & 29.67 & 30.00 & 26.67 & 20.00 & 32.00 \\
        ~~~-thinking & 14 & 42.26 & 17.24 & 13.00 & 25.25 & 1.62 & 52.67 & 22.67 & 75.00 & 66.20 & \cellcolor{lightyellow} 53.47 & 91.11 & 41.00 & 50.71 & 25.00 & 78.00 & 39.73 & 33.50 & 44.59 & 48.00 & 25.00 & 40.22 & 48.50 & 33.60 \\
        Qwen3-8B~\cite{yang2024qwen2} & 21 & 37.91 & 18.20 & 14.00 & 26.58 & 1.40 & 45.11 & 19.33 & 66.00 & 56.40 & 41.87 & 91.11 & 32.00 & 31.62 & 24.00 & 56.00 & 42.99 & 43.75 & 37.70 & 48.67 & 39.00 & 30.00 & 26.50 & 32.80 \\
        ~~~-thinking & 9 & 45.04 & 17.49 & 13.50 & 25.58 & 1.13 & 58.78 & 22.00 & 72.00 & 78.20 & 52.51 & 94.07 & 34.50 & 49.57 & 27.00 & 84.00 & 44.16 & 39.75 & 47.87 & 51.67 & 28.00 & 42.67 & 48.00 & 38.40 \\
        
        Qwen3-VL-4B~\cite{bai2025qwen2} & 19 & 38.67 & 18.81 & 12.50 & 28.57 & 2.04 & 47.44 & 17.00 & 86.00 & 58.00 & 45.81 & 94.07 & 34.50 & 37.32 & 26.00 & 60.00 & 38.19 & 37.00 & 37.38 & 42.00 & 34.00 & 35.56 & 34.00 & 36.80 \\
        ~~~-thinking & 11 & 43.70 & 15.81 & 15.50 & 21.26 & 0.00 & 58.00 & 22.00 & 80.00 & 75.20 & 51.79 & 92.59 & 37.50 & 46.72 & 28.00 & 82.00 & 39.91 & 29.00 & 44.26 & 55.67 & 23.00 & \cellcolor{lightyellow} 46.67 & 54.00 & 40.80 \\
        Qwen3-VL-8B~\cite{bai2025qwen2} & 16 & 42.10 & 25.74 & 11.50 & 43.52 & 0.69 & 45.78 & 20.67 & 81.00 & 53.80 & 48.44 & 92.59 & 28.50 & 47.01 & 24.00 & 68.00 & 43.53 & 41.75 & 43.28 & 51.00 & 29.00 & 41.33 & 45.00 & 38.40 \\
        ~~~-thinking & 7 & 45.66 & 20.97 & 16.00 & 31.23 & 0.00 & 59.11 & 27.00 & 77.00 & 74.80 & 52.99 & 94.81 & 37.50 & 52.14 & 28.00 & 58.00 & 43.62 & 30.50 & 49.84 & 60.00 & 28.00 & 43.11 & 51.50 & 36.40 \\
        Qwen3-VL-32B~\cite{bai2025qwen2} & 5 & 45.90 & 15.74 & 12.00 & 22.92 & 1.61 & 59.44 & 31.67 & 87.00 & 70.60 & 45.81 & 98.52 & 35.50 & 30.77 & 39.00 & 64.00 & 53.67 & 45.25 & 54.75 & 72.67 & 27.00 & 40.22 & 42.50 & 38.40 \\
        ~~~-thinking & 3 & \cellcolor{lightred} 51.06 & 16.34 & 13.00 & 23.92 & 0.20 & \cellcolor{lightred} 68.67 & 28.67 & 77.00 & 91.00 & \cellcolor{lightred} 59.45 & 96.30 & 39.00 & 59.54 & 40.00 & 80.00 & \cellcolor{lightred} 49.68 & 33.50 & 58.69 & 69.00 & 29.00 & \cellcolor{lightred} 50.00 & 54.00 & 46.80 \\

        \hline
        \textbf{Spatial Models} & & & & & & & & & & & & & & & & & & & & & & & & \\   
        Space-Qwen-3B~\cite{chen2024spatialvlm} & 33 & 27.26 & 16.35 & 21.00 & 17.28 & 4.24 & 36.33 & 11.33 & 71.00 & 44.40 & 27.75 & 44.44 & 22.50 & 25.64 & 24.00 & 26.00 & 29.77 & 27.00 & 38.36 & 28.85 & 18.00 & 16.22 & 19.00 & 14.00 \\
        SpaceThinker-3B~\cite{chen2024spatialvlm} & 30 & 33.83 & \cellcolor{lightred} 20.73 & 18.50 & 28.24 & 2.58 & 43.11 & 12.00 & 61.00 & 58.20 & 38.04 & 86.67 & 36.00 & 25.36 & 21.00 & 38.00 & 32.22 & 32.00 & 32.13 & 33.33 & 30.00 & \cellcolor{lightyellow} 28.89 & 16.00 & 39.20 \\
        Robobrain2.0-7B~\cite{team2025robobrain} & 24 & 35.52 & 18.41 & 16.50 & 25.58 & 0.62 & 36.78 & 17.67 & 67.00 & 42.20 & \cellcolor{lightyellow} 46.17 & 92.59 & 33.50 & 39.32 & 26.00 & 60.00 & \cellcolor{lightyellow} 41.36 & 40.50 & 40.66 & 46.67 & 31.00 & 21.78 & 23.00 & 20.80 \\
        SpaceR-7B~\cite{ouyang2025spacer} & 23 & \cellcolor{lightyellow} 36.42 & 19.40 & 12.50 & 27.91 & 7.60 & \cellcolor{lightyellow} 44.22 & 13.33 & 72.00 & 57.20 & 43.90 & 93.33 & 36.50 & 29.91 & 33.00 & 60.00 & 37.56 & 37.75 & 44.59 & 32.67 & 30.00 & 26.89 & 31.50 & 23.20 \\
        Cosmos-Reason2-8B~\cite{azzolini2025cosmos} & 15 & \cellcolor{lightred} 42.13 & \cellcolor{lightyellow} 20.59 & 14.50 & 31.23 & 0.76 & \cellcolor{lightred} 47.89 & 21.00 & 89.00 & 55.80 & \cellcolor{lightred} 50.00 & 92.59 & 33.00 & 49.86 & 23.00 & 58.00 & \cellcolor{lightred} 43.98 & 37.50 & 44.26 & 56.67 & 31.00 & \cellcolor{lightred} 40.22 & 49.00 & 33.20 \\
    \end{tabular}
    }
    \caption{\textbf{Performance of different models on SiT bench.} The highest and second-highest in each category are highlighted with \colorbox{lightred}{light red} and \colorbox{lightyellow} {light yellow}, respectively.}
    \label{tab:more_results}
    \vspace{-5mm}
\end{table*}

\onecolumn
\section{Prompt Construction, Sample Display and Test Results}
\label{appendixC}
\subsection{Global Perception \& Mapping}

\subsubsection{Scene Layout Reason }

\begin{tcolorbox}[prompt={Scene Layout Reason - CONSTRUCTION PROMPT},breakable]		
Task: Act as an objective visual observer. Provide a factual, spatially organized description of the scene in a single paragraph.\\
Context (Question to be answered later): \{user\_content\}\\
\newline
IMPORTANT: Do not attempt to answer the question or follow any multiple-choice format now. Your ONLY task is to provide a descriptive caption following these instructions:\\
Instructions:\\
1. Visual-Only Constraint (STRICT):\\
 - Describe ONLY what is directly visible in this specific image.\\
 - If an object mentioned in the Context is NOT visible in this frame, DO NOT mention it. Do not explain its absence, do not predict its location, and do not use "if/then" logic.\\
  - STRICTLY PROHIBITED: No spatial reasoning, no logical deductions, and no hints about the answer. Do not use phrases like "logically," "should be," "likely," or "based on the perspective."\\
2. Entity Hierarchy:\\
  - Use the Context ONLY to prioritize which VISIBLE objects to describe in detail.\\
  - Focus on visible objects, mainly about "\{visible\_objects\}". Explicitly state their orientation (e.g., "placed horizontally" or "placed vertically") and relative positions. Describe the objects you see using the same names as before.\\
3. Spatial Flow (Natural Narrative):\\
  - Follow a strict Left-to-Right scanning order from the camera's perspective.\\
  - Explicitly name the objects or architectural features at the Far Left and Far Right edges. State if they are "partially visible" or "cut by the frame."\\
4. Tone \& Format:\\
  - Use neutral, telegraphic prose. No flowery adjectives or overly detailed descriptions.\\
  - Output MUST be a single, continuous paragraph. Do not include any commentary, introductory remarks, or answers to the context question.
\end{tcolorbox}

\begin{tcolorbox}[prompt={Scene Layout Reason - DATA SAMPLE}]
\begin{minipage}[t]{0.25\linewidth}
  \centering
  \textbf{Ground Truth}\\
  D. Plush toy
\end{minipage}%
\hfill
\begin{minipage}[t]{0.7\linewidth}
  \textbf{Question}\\
  Based on these four images (image 1, 2, 3, and 4) showing the black waist bag from different viewpoints (front, left, back, and right), with each camera aligned with room walls and partially capturing the surroundings: If I am standing at the same spot and facing the same direction as shown in image 1, what is behind me?\\
  A. Window B. Black sofa C. Display shelves D. Plush toy
\end{minipage}
\vspace{0.5cm}

\begin{minipage}[t]{0.25\linewidth}
\vspace{0pt}
  \centering
  \includegraphics[width=\linewidth]{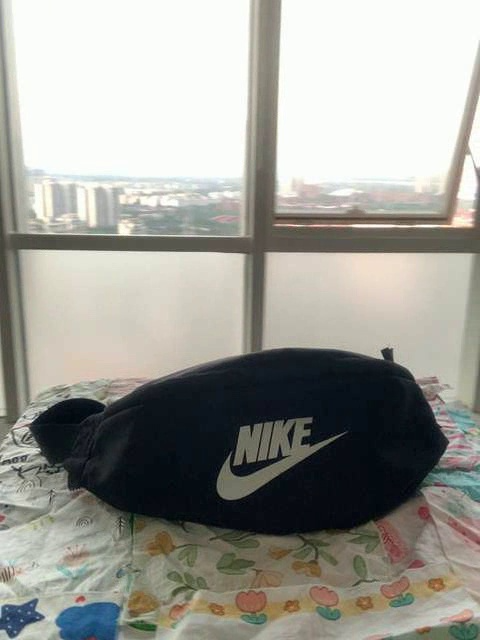}
\end{minipage}%
\hfill
\begin{minipage}[t]{0.7\linewidth}
  \textbf{Front}\\
  A black waist bag with a white Nike logo is placed horizontally on a patterned bedsheet in the foreground, centered within the frame. Behind the bag, a large window with vertical and horizontal frames spans the background, offering a view of distant buildings under an overcast sky; the window’s left and right edges are partially visible, cut by the frame.
\end{minipage}
\vspace{0.5cm}

\begin{minipage}[t]{0.25\linewidth}
\vspace{0pt}
  \centering
  \includegraphics[width=\linewidth]{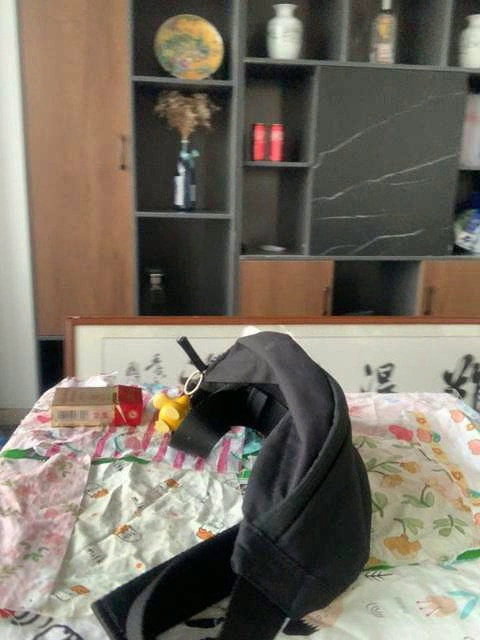}
\end{minipage}%
\hfill
\begin{minipage}[t]{0.7\linewidth}
  \textbf{Left}\\
  A black waist bag is placed vertically on a floral-patterned surface in the foreground, its strap and a small yellow plush toy attached to it visible. Behind the bag, a wooden-framed piece of calligraphy art is mounted on the wall. To the left of the frame, a tall wooden cabinet with a vertical handle is partially visible. To the right and extending across the background, dark gray display shelves with marble-like veining hold various objects: a circular decorative plate, a white vase, a bottle, two red cylindrical items, and other small containers. The far-right edge of the frame cuts off part of the display shelves and another white object.
\end{minipage}
\vspace{0.5cm}

\begin{minipage}[t]{0.25\linewidth}
\vspace{0pt}
  \centering
  \includegraphics[width=\linewidth]{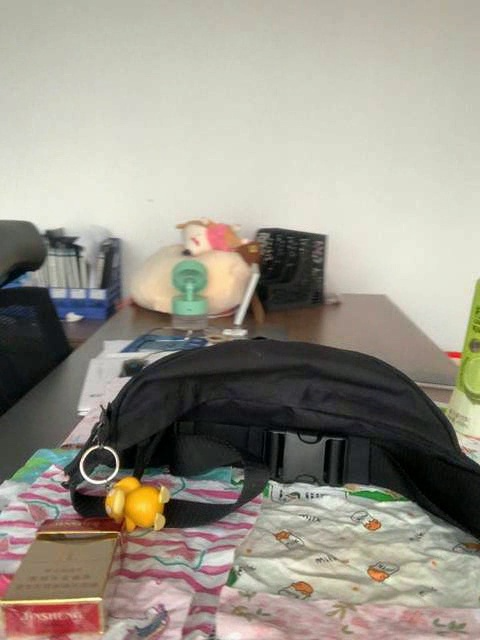}
\end{minipage}%
\hfill
\begin{minipage}[t]{0.7\linewidth}
  \textbf{Back}\\
  From the far left edge, a black office chair is partially visible, cut by the frame. Moving right, a dark wooden desk holds several items: a blue plastic file organizer with white binders stands vertically behind a light-colored plush toy with pink and brown accents, which is placed horizontally on the desk surface. To the right of the plush toy, a black rectangular object with vertical slats, possibly a small display shelf or speaker, sits upright. In the foreground, a black waist bag lies open on a patterned fabric surface, its strap extending toward the lower left where a small yellow plush keychain and a gold-colored box are positioned. The wall behind the desk is plain and off-white. On the far right edge, a green cylindrical container is partially visible, cut by the frame.
\end{minipage}
\vspace{0.5cm}

\begin{minipage}[t]{0.25\linewidth}
\vspace{0pt}
  \centering
  \includegraphics[width=\linewidth]{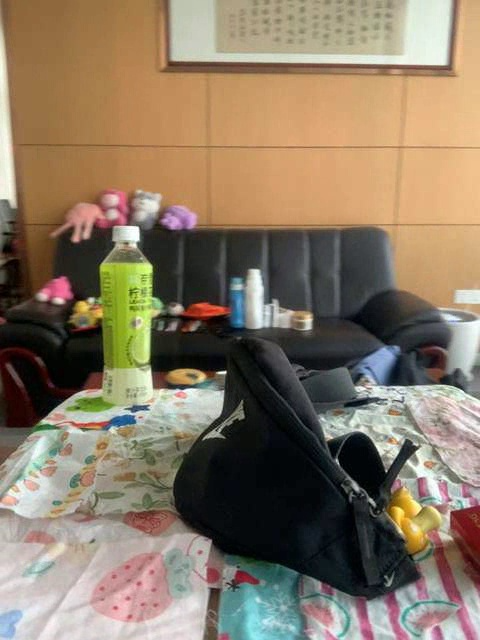}
\end{minipage}%
\hfill
\begin{minipage}[t]{0.7\linewidth}
  \textbf{Right}\\
  In the foreground, a black waist bag is placed vertically on a table covered with a colorful floral-patterned cloth; to its left stands a green-labeled beverage bottle placed vertically. Behind the table, a black sofa spans the midground, partially visible from left to right, with several plush toys resting on its backrest—pink, white, and purple ones are discernible. On the far left edge of the frame, a red chair is partially visible; on the far right, a white cylindrical object and part of a blue item are cut by the frame. Above the sofa, mounted on a wood-paneled wall, is a rectangular framed artwork with text, centered horizontally.
\end{minipage}
\end{tcolorbox}

\begin{tcolorbox}[prompt={Scene Layout Reason - TEST RESULTS},breakable]		
\textbf{Gemini3-pro}\\
Based on the descriptions provided for the four images:\\
1. Image 1 (Front View): The camera is facing the black waist bag and a window. Thus, the window is in front of the camera.\\
2. Image 3 (Back View): This camera viewpoint is the opposite of Image 1. It shows the bag from the other side, with a desk, a light-colored plush toy, and an off-white wall in its background.\\
3. Image 2 (Left View) and Image 4 (Right View): These show the bag from the remaining two sides, with backgrounds consisting of display shelves (Image 2) and a black sofa with plush toys (Image 4).\\
D. Plush toy\\

\textbf{Qwen3-8B thinking}\\
Looking at image 1: The black waist bag is in the foreground, centered. Behind it is a large window with vertical and horizontal frames, showing distant bulildings under an overcast sky. The window's edges are cut by the frame. Also, no black sofa, display shelves, or plush toys are visible here, So image 1 shows the front view where the waist bag is in front, and behind it is the window. So if I'm facing the same direction as image 1, then the window is behind me, But wait, the question is asking what's behind me when I'm facing the same direction as image 1. So if image 1 is the front view, then the window is behind the bag, so ehind me would be the window, But let me check other images to confirm.\\
...\\
But wait, the options also have B. Black sofa. Is there any possibility that the black sofa is behind the window? Let me check image 4. In image 4, the sofa is behind the table, but the table is in front of the soda. So if the person is in image 1's position, facing the same direction, then the window is directly behind. The sofa is not in image 1's view. So the answer should be A.\\
A. Window
\end{tcolorbox}

\subsubsection{Panoramic Counting}

\begin{tcolorbox}[prompt={Panoramic Counting - CONSTRUCTION PROMPT},breakable]		
Task: Act as a precise spatial annotator. Your goal is to provide a compressed, fluid, and spatially accurate English paragraph that describes the scene for 3D reasoning.\\
Question to be answered later: \{User Content\}\\
\newline
IMPORTANT: Do not attempt to count or answer any questions or following questions's format now. Your ONLY task is to provide a descriptive caption following these instructions:\\
Instructions:\\
1. Focus \& Filter (Entity Hierarchy):\\
  - Target Objects: For entities related to the question, describe their position and unique visual traits (e.g., "a silver sedan", "a red cement mixer with white stripes"). Do not use arbitrary IDs like 'Vehicle A' since you only see one view. Do not state the final count.\\
  - Spatial Anchors: For landmarks (buildings, fences), use only their functional name and color (e.g., "a red brick building", "a black metal fence"). STRICTLY IGNORE textures like "corrugated", "exposed rebar", or "wooden formwork".\\
  - Strict Exclusion: Skip weather, sky, and small debris.\\
2. Boundary Anchors (Stitching Logic):\\
  - Explicitly name the objects at the Far Left and Far Right edges. Clearly state if they are "partially cut" by the frame. This is the only way to align this view with others.\\
3. Spatial Flow (Natural Narrative):\\
  - Write as a single, continuous prose paragraph. Use transitional phrases like "Moving to the right," "Positioned behind this," or "Adjacent to."\\
  - Follow a strict Left-to-Right scanning order.\\
  - Use horizontal zones such as [Far Left, Center-Left, Center, Center-Right, Far Right].\\
4. Constraint:\\
  - Avoid flowery adjectives. Use "Telegraphic yet Fluent" prose.\\
  - DO NOT use bullet points, brackets like [Extreme Left], or artificial IDs like "Vehicle A".\\
  - Use specific visual identifiers (brand, color, or relative size) that would allow another person to recognize the same object in a different image.
\end{tcolorbox}

\begin{tcolorbox}[prompt={Panoramic Counting - DATA SAMPLE}]
\begin{minipage}[t]{0.25\linewidth}
  \centering
  \textbf{Ground Truth}\\
  4
\end{minipage}%
\hfill
\begin{minipage}[t]{0.7\linewidth}
  \textbf{Question}\\
  How many vehicles are visible in this scene?
\end{minipage}
\vspace{0.5cm}

\begin{minipage}[t]{0.25\linewidth}
\vspace{0pt}
  \centering
  \includegraphics[width=\linewidth]{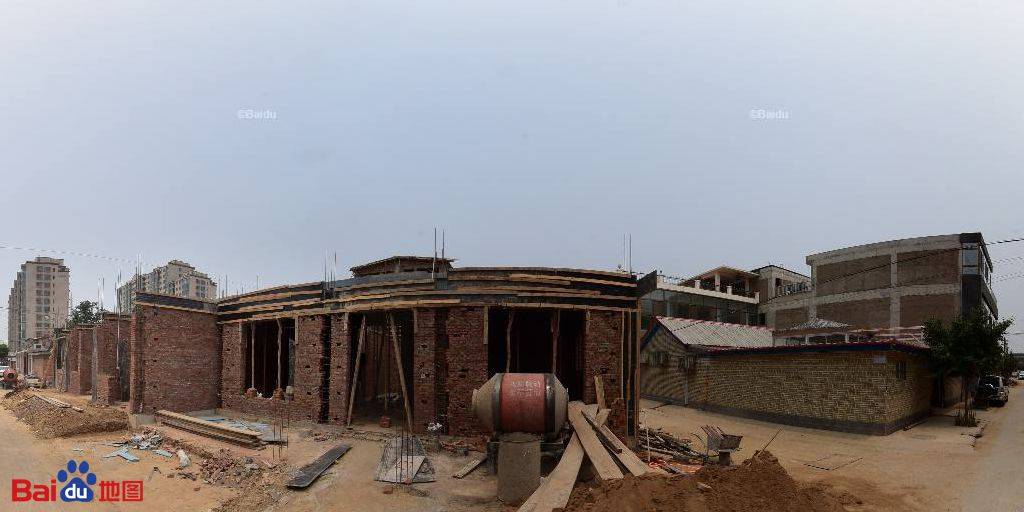}
\end{minipage}%
\hfill
\begin{minipage}[t]{0.7\linewidth}
  \textbf{North}\\
  At the far left edge, a partially cut tall beige apartment building stands behind a red brick structure under construction. Moving right, the center-left features a long red brick building with open window frames and wooden supports, fronted by a red cement mixer on a concrete base. Positioned behind this, in the center-right, is another red brick building with a flat roof and visible upper-floor windows. Adjacent to it, at the far right edge, a low red brick wall with a dark gray roofline runs horizontally, partially obscuring a black sedan parked behind it.
\end{minipage}
\vspace{0.5cm}

\begin{minipage}[t]{0.25\linewidth}
\vspace{0pt}
  \centering
  \includegraphics[width=\linewidth]{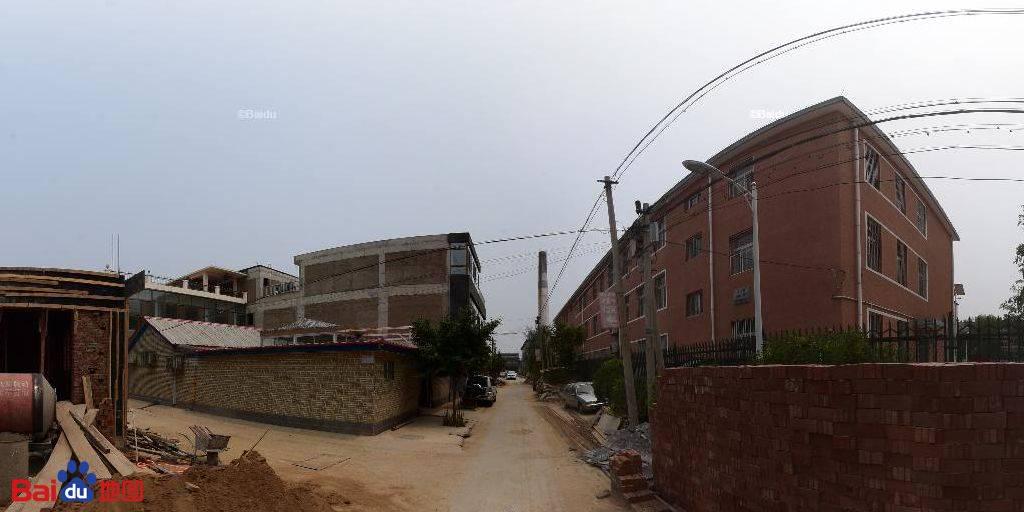}
\end{minipage}%
\hfill
\begin{minipage}[t]{0.7\linewidth}
  \textbf{East}\\
  At the far left edge, a partially cut red brick structure with an open front and scattered construction materials is visible. Moving to the right, a low beige brick wall runs parallel to a dirt road, with a small tree and a dark-colored vehicle parked beside it. Further along the center, another dark vehicle is parked near a taller unfinished gray building with exposed structural elements. Positioned behind this, a silver sedan is visible further down the road. To the center-right, a tall red brick building with multiple windows and white trim stands adjacent to a black metal fence. At the far right edge, a red brick wall runs vertically, partially obscuring the base of the red brick building and extending out of frame.
\end{minipage}
\vspace{0.5cm}

\begin{minipage}[t]{0.25\linewidth}
\vspace{0pt}
  \centering
  \includegraphics[width=\linewidth]{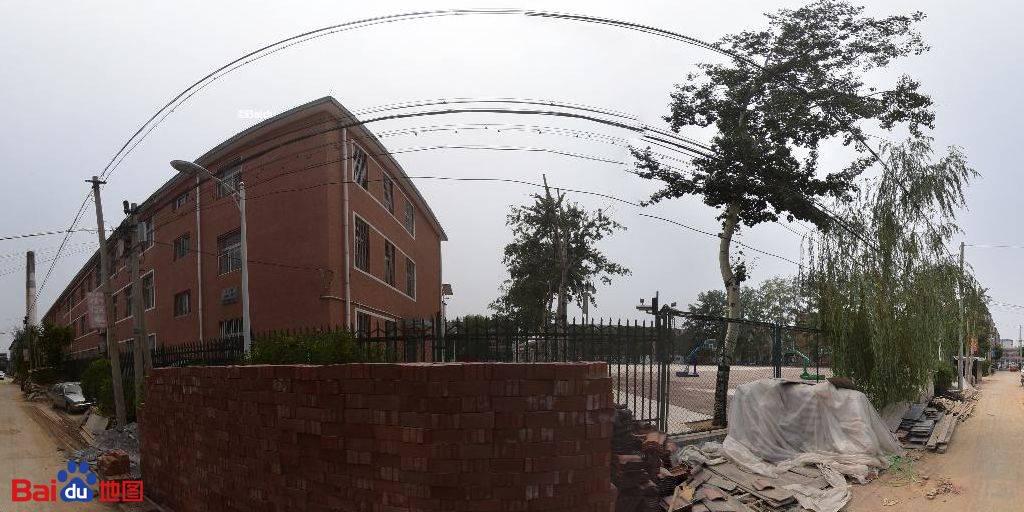}
\end{minipage}%
\hfill
\begin{minipage}[t]{0.7\linewidth}
  \textbf{South}\\
  At the far left edge, a silver sedan is partially cut by the frame, parked beside a red brick building that extends toward the center. Moving to the right, a black metal fence runs parallel to the building, separating it from an open paved area with playground equipment visible behind it. Adjacent to the fence on the right, a large pile of construction materials covered by a gray tarp sits near a tall weeping willow tree. Further right, the scene continues along a dirt road lined with utility poles and sparse vegetation, leading to a distant structure at the far right edge, also partially cut by the frame.
\end{minipage}
\vspace{0.5cm}

\begin{minipage}[t]{0.25\linewidth}
\vspace{0pt}
  \centering
  \includegraphics[width=\linewidth]{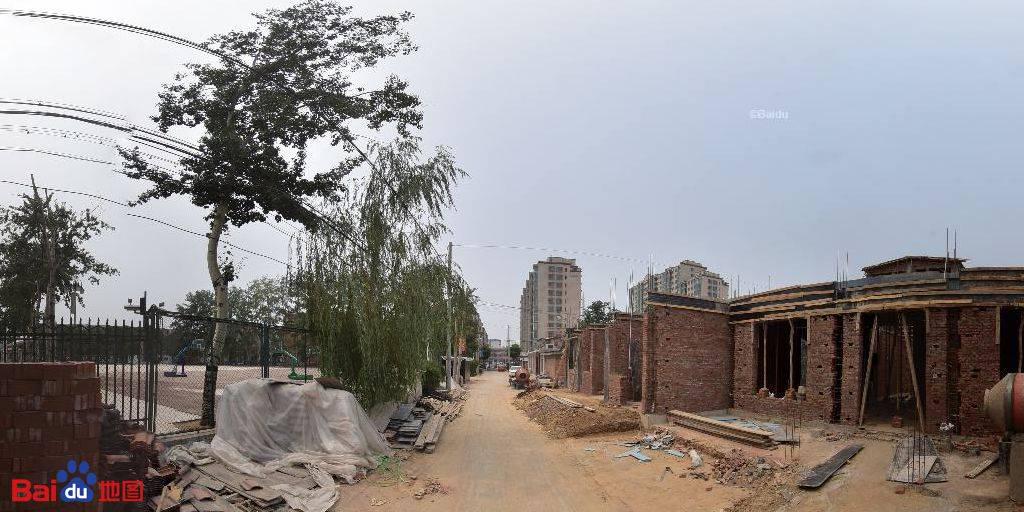}
\end{minipage}%
\hfill
\begin{minipage}[t]{0.7\linewidth}
  \textbf{West}\\
  At the far left edge, a black metal fence runs vertically, partially cut by the frame, with a stack of red bricks and a large tarp-covered pile in front. Moving to the right, a tall tree with drooping green leaves stands beside the fence, its branches extending over a dirt path. Further along the center-left, another leafy tree is visible behind the fence line. In the center distance, a silver sedan is parked on the dirt road, facing away from the viewer. Positioned behind this vehicle, two multi-story residential buildings rise against the horizon. To the center-right, a row of red brick buildings under construction extends along the right side of the path, their open frames revealing interior supports. At the far right edge, a red cement mixer with white stripes is partially cut by the frame, positioned adjacent to the last brick structure.
\end{minipage}
\end{tcolorbox}

\begin{tcolorbox}[prompt={Panoramic Counting - TEST RESULTS},breakable]		
\textbf{Gemini3-pro}\\
1.Vehicle 1 (Black/Dark Sedan):\\
In Image 1(North),at the far right edge,a black sedan is parked behind a low red brick wall.\\
In Image 2 (East,at the far left/center-left,a dark-colored vehicle is parked beside a low beige brick wall.\\
Deduction: Due to the overlapping nature of the images (Right edge of North overlaps Left edge of East),these are the same vehicle located in the Northeast corner.\\
2.Vehicle 2 (Dark Vehicle near Gray Building):\\
In Image 2 (East),further along the center,the caption explicitly states "another dark vehicle" is parked near a taller unfinished gray building.\\
Deduction: The use of"another" and the central location in the East view distinguishes this from the first vehicle.\\
3.Vehicle 3 (Silver Sedan-East/South):\\
In Image 2 (East),to the center-right/distance,a silver sedan is visible further down the road. In Image 3(South),at the far left edge,a silver sedan is partially cut by the frame.\\
Deduction: The Right edge of the East view overlaps with the Left edge of the South view.These represent the same silver sedan located in the Southeast direction.\\
4.Vehicle 4 (Silver Sedan-West):\\
In Image 4 (West),in the center distance,a silver sedan is parked on the dirt road facing away.\\
Deduction: This vehicle is located to the West.It is distinct from Vehicle 3 (which is to the Southeast) and Vehicle 1/2 (which are to the Northeast/East).\\
Note: The "red cement mixer" mentioned in Images 1 and 4 is described as being "on a concrete base," implying it is stationary construction equipment rather than a vehicle.\\
Total Count:\\
Vehicle 1: Black/Dark sedan (Northeast) Vehicle 2:Dark vehicle (East)
.Vehicle 3: Silver sedan (Southeast) Vehicle 4: Silver sedan (West)\\
Answer: 4\\

\textbf{Qwen3-235B-A3B thinking}\\
...\\
4.Silver Sedan 2 (West):\\
Image 4 (West): Silver sedan parked on the dirt road (center distance).\\
Conclusion: Unique to the West side of the scene. No overlap with other images.Count as 1.\\
5.Dark Vehicle (East):\\
Image2 (East): A second dark vehicle near an unfinished gray building.\\
Verification: No matching description in other images (e.g, Image 3 mentions playground equipment and construction piles but no additional dark vehicle).This is a unique vehicle located East of the observer. Count as 1.\\
Final Count:\\
Cement mixer: 1\\
Black sedan: 1\\
Silver sedan 1 (Southeast): 1\\
Silver sedan 2 (West): 1\\
Dark vehicle (East): 1\\
Total unique vehicles: 5\\
Your answer: 5
\end{tcolorbox}

\subsection{Navigation \& Planning}

\subsubsection{Path Planning Logic}

\begin{tcolorbox}[prompt={Path Planning Logic - Construction Prompt},breakable]		
Task: You are an expert spatial navigator and planner. I will provide you with a floor plan image and a text description of the room layout (names, coordinates, and connections).\\
Your goal is to create a multiple-choice question (MCQ) that tests a model's ability to plan a route or understand spatial relationships in this 2D plane.\\
\newline
Input Information:\\
1. Floor Plan Image: Visual representation of the layout.\\
2. Room Layout Info: A list of rooms with their bounding box coordinates/vertices, centroids, and graph connectivity. Note that the coordinates correspond to the image pixels.\\
\newline
Requirements for Question Generation:\\
1. Scenario: Create a navigation task, e.g., "How do I get from the [Start Room] to the [End Room]?" or "What is the path from [Room A] to [Room B]?".\\
2. Reasoning: You MUST rely on the visual map to determine valid connections (doors, openings) and specific movements (turns, directions). The provided text info helps you locate rooms, but the image is the ground truth for navigation.\\
3. Output Format:\\ 
- Caption: A descriptive caption of the floor plan. \\
\noindent\hspace*{2em}- You MUST describe the general layout and relative positions of rooms based \noindent\hspace*{2em} on the image (e.g., "The front door is located in the bottom-left corner. \noindent\hspace*{2em} The living room occupies the central area. The kitchen is adjacent to the \noindent\hspace*{2em} living room on the north side."). \\
\noindent\hspace*{2em}- Do NOT reveal the specific answer to the question in this caption.\\
- Question: A clear question describing the navigation task.\\
- Options: Provide 4 options (A, B, C, D).\\
\noindent\hspace*{2em}- One correct option describing the valid path.\\
\noindent\hspace*{2em}- Three plausible but incorrect options (distractors).\\
\noindent\hspace*{2em}- CRITICAL: The options MUST be detailed and describe physical movements, \noindent\hspace*{2em} not just room transitions. Use terms like "walk straight", "turn left", \noindent\hspace*{2em} "turn right", "pass through the door on the left", etc. (e.g., "Enter \noindent\hspace*{2em} through the front\_door\_0, turn left to cross the living\_0, then turn right \noindent\hspace*{2em} to enter the kitchen\_0 door").\\
- Answer: The correct option label (A, B, C, or D).\\
- Explanation: A brief explanation of why the correct path is valid and others are not based on the visual evidence.\\
\newline
Room Layout Info provided:\\
\{layout\_info\}\\
\newline
Please generate the response in the following JSON format ONLY, without markdown code blocks:\\
\{\\
\noindent\hspace*{2em}  "caption": "...",\\
\noindent\hspace*{2em}  "question": "...",\\
\noindent\hspace*{2em}  "options": \{\\
\noindent\hspace*{4em}    "A": "...",\\
\noindent\hspace*{4em}    "B": "...",\\
\noindent\hspace*{4em}    "C": "...",\\
\noindent\hspace*{4em}    "D": "..."\\
\noindent\hspace*{2em}  \},\\
\noindent\hspace*{2em}  "answer": "A",\\
\noindent\hspace*{2em}  "explanation": "..."\\
\}
\end{tcolorbox}

\vspace{0.5cm}

\vspace{0.5cm}


\subsubsection{Ego/Objects-motion Perception}

\begin{tcolorbox}[prompt={Ego/Objects-motion Perception - DATA SAMPLE}]
\begin{minipage}[t]{0.25\linewidth}
  \centering
  \textbf{Ground Truth}\\
  A
\end{minipage}%
\hfill
\begin{minipage}[t]{0.7\linewidth}
  \textbf{Question}\\
  The front view corresponds to the north direction. If the dark maroon suv parked near construction barricades in the back view moves 5 meters north while all other objects remain stationary, does the dark maroon suv parked near construction barricades in the back view get closer to the white suv waiting near the intersection in the back right view?\\
  \textbf{Options}\\
  A.yes B.no
\end{minipage}
\vspace{0.5cm}

\begin{minipage}[t]{0.25\linewidth}
\vspace{0pt}
  \centering
  \includegraphics[width=\linewidth]{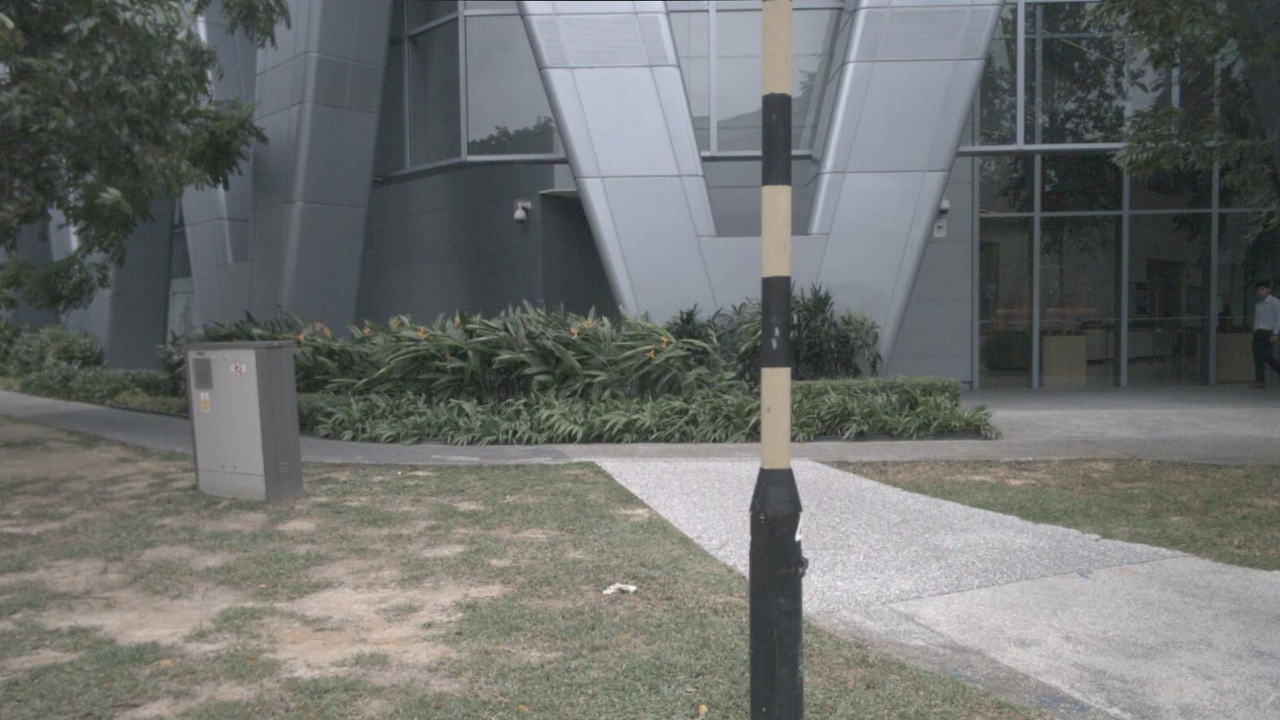}
\end{minipage}%
\hfill
\begin{minipage}[t]{0.7\linewidth}
  \textbf{Back\_Left}\\
  A gray utility box stands on a patch of dry grass at the far left, partially cut by the frame. Moving to the right, a dense row of green shrubs with yellow flowers runs horizontally across the midground, positioned in front of a modern gray building with large glass panels and angular metallic cladding. Centered in the foreground is a vertical pole painted with alternating black and beige segments, standing upright on the grass near a concrete pathway that angles diagonally toward the right. The building continues behind the shrubs, its reflective windows and structural folds extending to the far right edge, where a glass entrance with visible interior elements is partially cut by the frame. A person in light clothing appears faintly near the entrance on the far right, standing still.
\end{minipage}
\vspace{0.5cm}

\begin{minipage}[t]{0.25\linewidth}
\vspace{0pt}
  \centering
  \includegraphics[width=\linewidth]{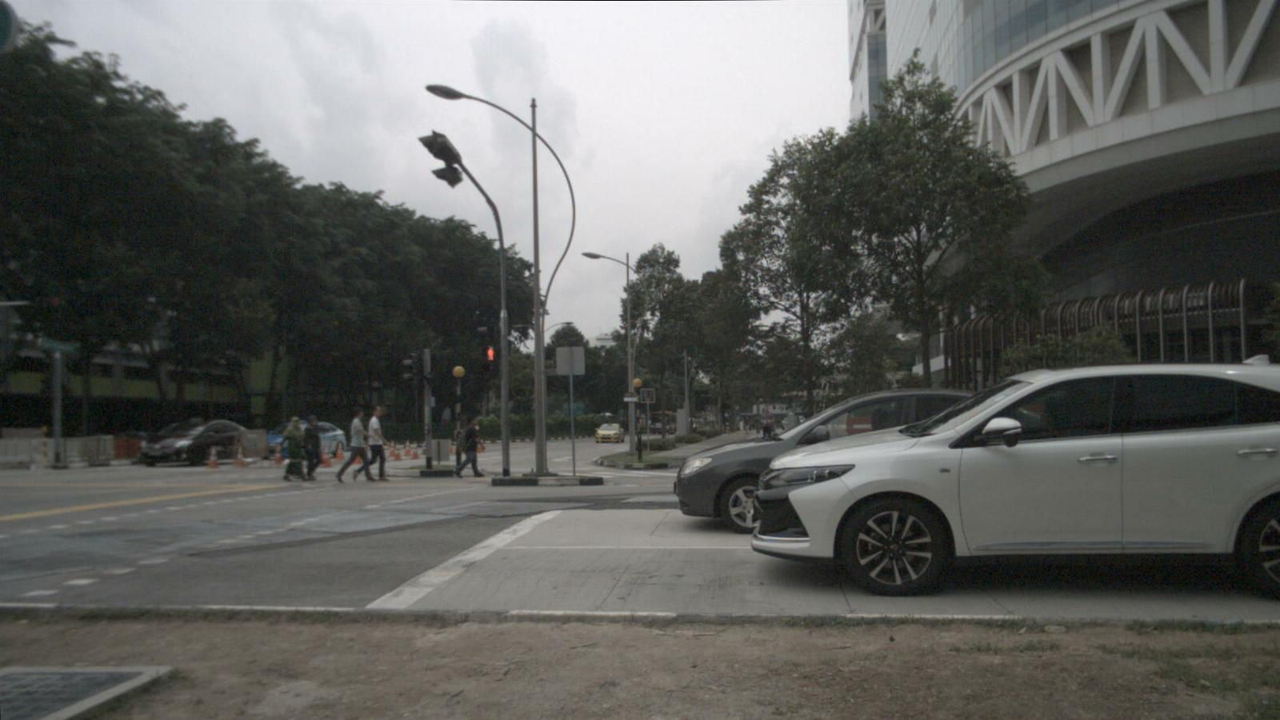}
\end{minipage}%
\hfill
\begin{minipage}[t]{0.7\linewidth}
  \textbf{Back\_Right}\\
  A dark maroon SUV is parked near construction barricades on the far left, partially cut by the frame, with a group of pedestrians crossing the road ahead. Moving to the right, a white SUV waits near the intersection in the center-right, positioned just behind a gray sedan also stopped at the junction. Adjacent to the white SUV on the far right is a modern building with curved white architectural elements, its facade partially visible and framing the scene. The white SUV remains stationary while the maroon SUV, if moved five meters north, would shift closer toward the intersection zone occupied by the white SUV, reducing the lateral distance between them within the same traffic lane alignment.
\end{minipage}
\vspace{0.5cm}

\begin{minipage}[t]{0.25\linewidth}
\vspace{0pt}
  \centering
  \includegraphics[width=\linewidth]{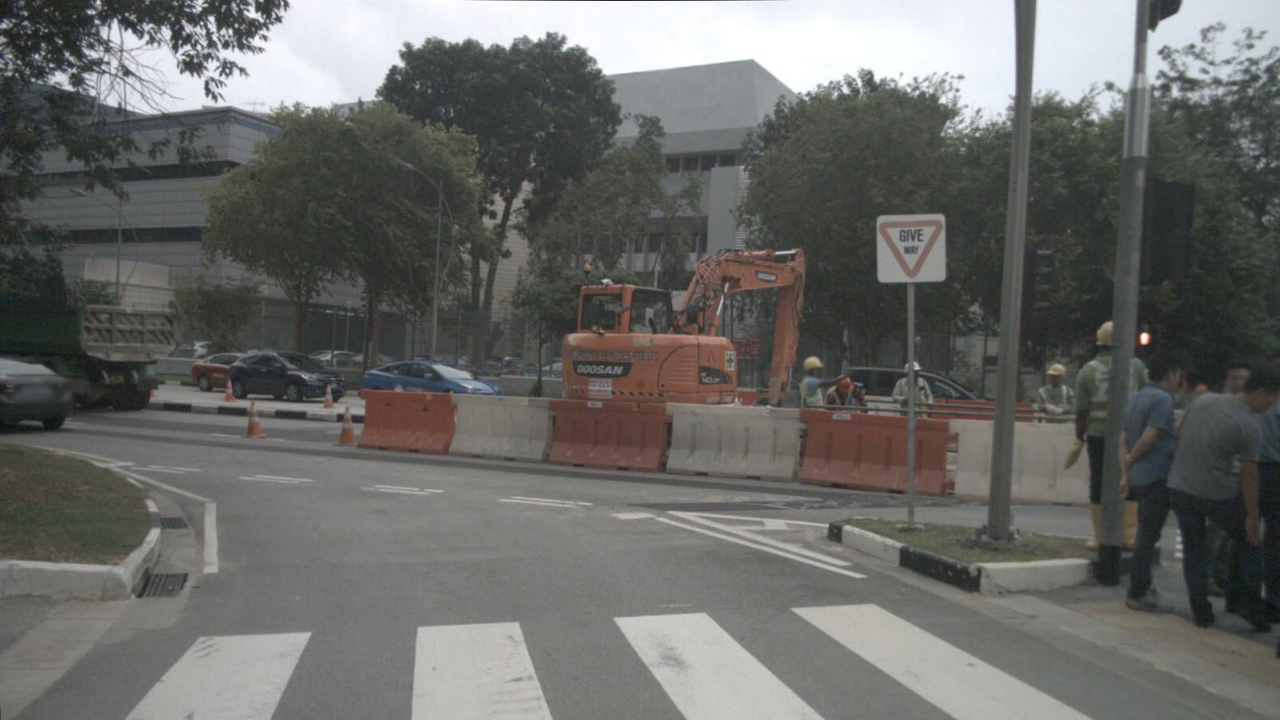}
\end{minipage}%
\hfill
\begin{minipage}[t]{0.7\linewidth}
  \textbf{Front}\\
  A dark maroon SUV is parked near orange construction barricades on the left side of the road, partially cut by the far left edge of the frame. Moving to the right, a blue sedan is stopped behind the barricades, with two orange traffic cones in front of it. Centered in the scene is an orange Doosan excavator positioned behind a line of white and orange concrete barriers. Adjacent to the excavator on the right, a white SUV waits near the intersection, partially visible at the far right edge of the frame. Behind the excavator and slightly to its right, a group of workers in yellow helmets and high-visibility vests stand near additional barricades. A triangular “GIVE WAY” sign mounted on a gray pole stands prominently on the far right, just before the white SUV. The background features a large light-gray building surrounded by green trees, forming a consistent backdrop across the center and right portions of the view.
\end{minipage}
\vspace{0.5cm}

\begin{minipage}[t]{0.25\linewidth}
\vspace{0pt}
  \centering
  \includegraphics[width=\linewidth]{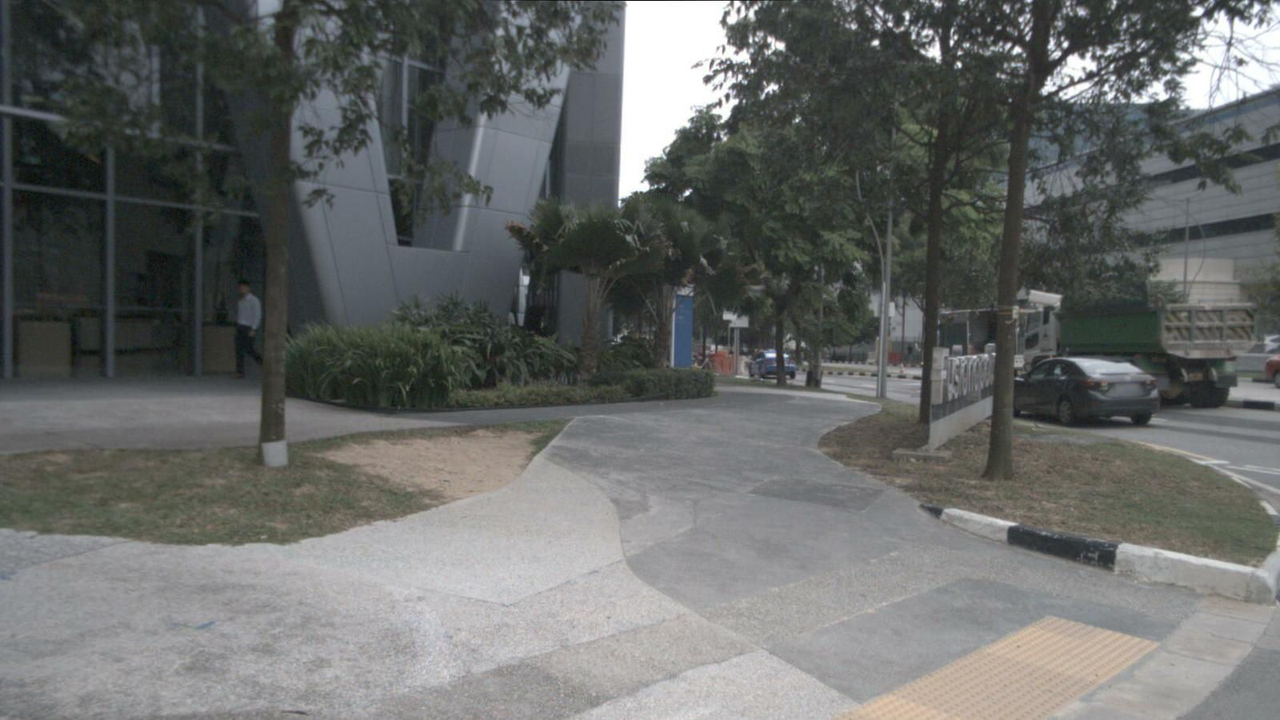}
\end{minipage}%
\hfill
\begin{minipage}[t]{0.7\linewidth}
  \textbf{Front\_Left}\\
  A modern gray building with large glass panels occupies the far left, partially cut by the frame, with a person standing near its entrance. Moving to the right, a row of lush green shrubs lines the edge of a paved walkway that curves gently toward the center. A tall tree with dense foliage stands beside the path, followed by another tree slightly further right. Positioned behind this second tree is a white sign mounted on a low concrete wall. Further right, a dark gray sedan is parked along the curb, adjacent to a large green dump truck with a visible cab and open bed, also partially cut by the far right edge of the frame. The road surface extends into the background where additional vehicles are faintly visible near distant construction barricades.
\end{minipage}
\vspace{0.5cm}

\begin{minipage}[t]{0.25\linewidth}
\vspace{0pt}
  \centering
  \includegraphics[width=\linewidth]{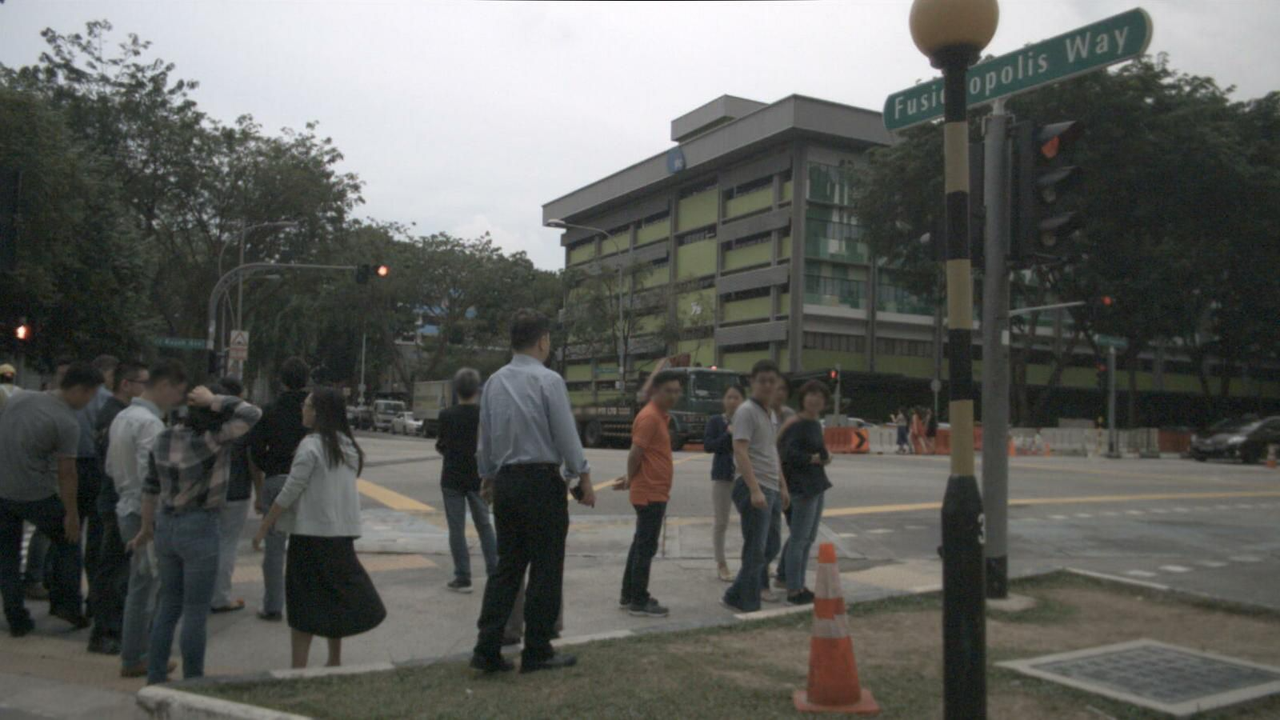}
\end{minipage}%
\hfill
\begin{minipage}[t]{0.7\linewidth}
  \textbf{Front\_Right}\\
  A group of pedestrians stands on the sidewalk near a crosswalk, with several individuals wearing casual attire including a man in a light blue shirt and dark trousers facing away from the camera, a woman in a white top and black skirt, and others in plaid shirts or orange tops. Moving to the right, a green multi-story building with large windows occupies the background, positioned behind the crowd. Adjacent to the building, a traffic signal pole with a street sign reading “Fusionopolis Way” rises vertically, partially obscuring the view of the road beyond. On the far right edge, a dark maroon SUV is parked near orange construction barricades, partially cut by the frame. Further right, beyond the barricades, a white SUV waits near the intersection, also partially visible. A single orange traffic cone sits on the grass verge in the foreground near the curb. The scene is framed by trees on the far left, which extend into the background, while the road surface stretches horizontally across the lower portion of the image.
\end{minipage}

\end{tcolorbox}

\begin{tcolorbox}[prompt={Ego/Objects-motion Perception - TEST RESULTS},breakable]		
\textbf{Gemini3-pro}\\
Based on the provided captions and scene analysis:\\
1. Orientation: The Front view corresponds to the North direction. Consequently, the Back
view corresponds to the South direction.\\
2. Object Positions:
Dark Maroon SUV: It is identified in the Back view (looking South), the Back Right view, and the Front view. In the Back view, it is parked beside construction
barricades. Being visible in the Back view implies it is positioned towards the South or alongside the ego vehicle.\\
White SUV: It is identified in the Front view, Front Right view, and Back Right view,waiting near an intersection. Since it is waiting at an intersection visible in the Front view (North),it is located North of the ego vehicle.\\
3. Movement Analysis:
The question states the dark maroon SUV moves 5 meters North.\\
Since the white SUV is located at the intersection to the North, and the maroon SUV moves North (towards the intersection), the maroon SUV is closing the distance to the white SUV.\\
4. Caption Evidence: The caption for the Back\_Right view explicitly contains a scenario description: "\textit{The white SUV remains stationary while the maroon SUV, if moved five meters north, would shift closer toward the intersection zone occupied by the white SUV reducing the lateral distance between them within the same traffic lane alignment.}" This text directly confirms the outcome.\\ 
Answer: A.yes\\

\textbf{Qwen3-235b-a22b-thinking}\\
Why the Answer is "Yes":\\
The maroon SUV starts south of the ego car, and the white SUV is north of the ego car. Moving north directly closes the gap between them.\\
No other objects or road geometry (e.g., curves, barriers) contradict this, as the captions describe a consistent urban road layout with the intersection as a central reference point.\\
The question specifically asks about the white SUV "waiting near the intersection
in the Back\_Right view," which is the same vehicle visible in the Front view,confirming its position north of the ego car.\\
Conclusion: After moving 5 meters north, the dark maroon SUV gets closer to the white SUV.\\
Answer: A.yes
\end{tcolorbox}

\subsection{Multi-View \& Geometric Reasoning}

\subsubsection{View Consistency}

\begin{tcolorbox}[prompt={View Consistency - CONSTRUCTION PROMPT},breakable]		
Task: Act as an objective visual observer. Provide a factual, spatially organized description of the scene in a single paragraph.\\
Context (Question to be answered later): {user\_content}\\
IMPORTANT: Do not attempt to answer the question or follow any multiple-choice format now. Your ONLY task is to provide a descriptive caption following these instructions:\\
\newline
Instructions:\\
1. Visual-Only Constraint (STRICT):\\
- Describe ONLY what is directly visible in this specific image. \\
- If an object mentioned in the Context is NOT visible in this frame, DO NOT mention it. Do not explain its absence, do not predict its location, and do not use "if/then" logic.\\
- STRICTLY PROHIBITED: No spatial reasoning, no logical deductions, and no hints about the answer. Do not use phrases like "logically," "should be," "likely," or "based on the perspective."\\
2. Entity Hierarchy:\\
- Use the Context ONLY to prioritize which VISIBLE objects to describe in detail.\\
- Focus on visible furniture, structural elements (walls, corners, floor), and their relative positions.\\
3. Spatial Flow (Natural Narrative):\\
- Follow a strict Left-to-Right scanning order from the camera's perspective.\\
- Explicitly name the objects or architectural features at the Far Left and Far Right edges. State if they are "partially visible" or "cut by the frame."\\
4. Tone \& Format:\\
- Use neutral, telegraphic prose. No flowery adjectives.\\
- Output MUST be a single, continuous paragraph. Do not include any commentary, introductory remarks, or answers to the context question.
\end{tcolorbox}

\begin{tcolorbox}[prompt={View Consistency - DATA SAMPLE}]
\begin{minipage}[t]{0.25\linewidth}
  \centering
  \textbf{Ground Truth}\\
  C. left
\end{minipage}%
\hfill
\begin{minipage}[t]{0.7\linewidth}
  \textbf{Question}\\
  Standing at desk, gazing at shelves, where should whiteboard be?\\
  A. Back B. Front C. left D. right
\end{minipage}
\vspace{0.5cm}

\begin{minipage}[t]{0.25\linewidth}
\vspace{0pt}
  \centering
  \includegraphics[width=\linewidth]{pic/ViewSpatial1.png}
\end{minipage}%
\hfill
\begin{minipage}[t]{0.7\linewidth}
  \textbf{Front}\\
  On the far left, a portion of an orange upholstered chair and a rolled-up cylindrical object rest against a white wall panel that extends vertically; adjacent to this, a black cabinet holds a blue spray bottle on its top surface, with a soccer ball and a deflated pump lying on the striped purple carpet at its base. Moving right, two black office chairs with wheeled bases sit before a long wooden desk against the back wall, which supports a dark computer monitor and keyboard; above the desk, a bulletin board displays various papers and small framed images. To the far right, a built-in wooden shelving unit with multiple compartments contains scattered items including bottles and papers, its structure extending toward the edge of the frame where another partial black chair leg is visible.
\end{minipage}
\vspace{0.5cm}

\begin{minipage}[t]{0.25\linewidth}
\vspace{0pt}
  \centering
  \includegraphics[width=\linewidth]{pic/ViewSpatial2.png}
\end{minipage}%
\hfill
\begin{minipage}[t]{0.7\linewidth}
  \textbf{Left}\\
  On the far left, a white wall corner is partially visible, adjacent to a large whiteboard covered in multicolored writing and diagrams, mounted on the wall. Below the whiteboard sits a black mini-fridge with a blue spray bottle and a dark plate on its top surface. To the right of the fridge, an office chair with a grey backrest and seat is positioned on a striped carpet, facing a light wood desk that extends across the mid-right portion of the frame. On the desk’s far right end, a computer monitor is partially visible, cut by the frame edge, with some small items scattered near its base. The wall behind the desk has a beige upper section and a darker lower panel, with papers or posters affixed above the desk surface.
\end{minipage}

\end{tcolorbox}

\begin{tcolorbox}[prompt={View Consistency - TEST RESULTS},breakable]		
\textbf{Gemini3-pro}\\
Based on the descriptions provided:\\
1.Image 1 establishes the layout from left to right: An orange chair and black cabinet (far left) → Desk with computer (center/right) → Shelves (far right).\\
2. Image 2 confirms the whiteboard is located above the black cabinet/mini-fridge, which is to the left of the desk.\\
If you are standing near the shelves (which are on the far right) and gazing at the desk (which is in the center), the whiteboard (located above the cabinet on the far left) would be to the left of the desk and to your left.\\
Answer: C\\

\textbf{Qwen3-30b-a3b-instruct}\\
(no reasoning)
A. back\\

\textbf{Qwen3-30b-a3b-thinking}\\
Let's make sure we are not confusing the direction. If we are standing at the right side of the room (near the shelves) and facing the desk (which is in front of us, along the back wall), then:\\
The desk is in front of us.\\
The left wall (with the whiteboard) is to our left.\\
The right wall would be behind us? But we are standing near the right wall (because the shelves are at the far right, so we are at the right wall). so the right wall is behind us? Actually, we are standing at the right wall, so the right wall is to our right? Wait, no: if we are standing at the right wall (meaning we are near the right wall, facing the desk which is in front of us), then the right wall is to our right? Actually, if we are facing the desk (which is in front of us). then:\\
The direction we are facing: forward (toward the desk).\\
Our left side: toward the left wall.\\
Our right side: toward the right wall(but we are standing at the right wall,so the right wall is immediately to our right? Actually, we are standing at the right wall, so the right wall is behind us? Or we are standing in the room, so the right wall is to our right?).\\
...\\
C
\end{tcolorbox}

\subsubsection{Perspective Shift}

\begin{tcolorbox}[prompt={Perspective Shift - CONSTRUCTION PROMPT},breakable]		
Task: Act as a 3D Spatial Intelligence Architect. Your goal is to generate a high-fidelity spatial reasoning benchmark item based on the provided image.\\
Task Objective: Synthesize a fluid textual description of the scene and a complex multiple-choice question that necessitates 3D mental modeling (e.g., perspective transformation or mental rotation).\\
Step 1: Spatial Encoding (Internal Logic) Before writing, internally establish a Right-Handed Coordinate System:\\
- Origin (0,0,0): The observer's current position.\\
- +Y Axis: The observer's initial forward vector.\\
- +X Axis: To the observer's right.\\
- +Z Axis: Upward.\\
Step 2: Descriptive Prose (The Caption) Write a single, fluid English paragraph (no bullets).\\
- Entity Grounding: Identify key objects using [Color + Functional Name].\\
- Spatial Anchoring: Define the observer's initial orientation clearly. Use the established coordinate logic to describe positions (e.g., "to your front-left," "positioned 5 meters ahead on your right flank").\\
- Orientation: Specify which way target objects (vehicles, etc.) are facing.\\
- Boundary Anchors: Note objects at the [Far-Left] and [Far-Right] edges as stitching anchors.\\
- Noise Filtering: Ignore weather, textures, and watermarks.\\
Step 3: Question Engineering (The Challenge) Create a multiple-choice question of one of these types:\\
1. Mental Rotation: A 90/180/270 degree turn by the observer.\\
2. Perspective Switching: Reasoning from the POV of another object/entity in the scene.\\
3. Geometric Prediction: Determining visibility or collision after a specific movement. Requirement: The question must require a 3D mental map. Simple 2D keyword matching must fail.\\
Step 4: Output Format (Strict JSON) Output ONLY a JSON object with these keys:\\
- caption: The fluid descriptive paragraph.\\
- question\_type: Choose from ["Mental Rotation", "Perspective Shift", "Spatial Navigation"].\\
- question: The multiple-choice question with four options (A, B, C, D).\\
- options: \{"A": "...", "B": "...", "C": "...", "D": "..."\}\\
- answer: The correct option letter.\\
- derivation: A rigorous geometric proof. Must include simplified relative coordinates (e.g., "Initial: Object at (+1, 2); After 180° rotation: Object at (-1, -2)") to justify the answer.
\end{tcolorbox}




\subsubsection{Pure Mental Rotation}

\begin{tcolorbox}[prompt={Pure Mental Rotation - CONSTRUCTION PROMPT},breakable]		
Task: Act as a precise LEGO spatial annotator. Your goal is to provide a compressed, fluid, and spatially accurate English paragraph that describes the visible geometry of the scene.\\
Question to be answered later: {user\_content}\\
IMPORTANT - STRICT CONSTRAINTS:\\
1. NO ANALYSIS OR INFERENCE: You must ONLY describe what is visually present. DO NOT use words like "suggesting," "indicating," "implies," "perspective," "angle," or "view." DO NOT summarize the scene's orientation at the end.\\
2. NO SPOILERS: Do not explicitly state if the view is top-down, side, or front. Just describe the exposed surfaces (studs vs. sides).\\
3. NO COUNTING: Do not count objects. Focus on their spatial relationships.\\
\newline
Instructions for Description:\\
1. Spatial Flow (Natural Narrative):\\
  - Write as a single, continuous prose paragraph.\\
  - Follow a strict Left-to-Right scanning order. Start with the object on the far left and move across the scene to the far right.\\
  - Use transitional phrases to link objects, such as "Moving to the right," "Adjacent to this," "Positioned behind the green tree," or "To the immediate right."\\
2. LEGO Geometry \& Visual Facts (The "What", not the "Why"):\\
  - Describe the visible surfaces strictly as visual facts:\\
\noindent\hspace*{2em}    - Studs vs. Sides: State clearly if you see the grid of circular studs on \noindent\hspace*{2em} top surfaces OR the smooth vertical sides of bricks. (e.g., "The red roof \noindent\hspace*{2em} shows a grid of studs" vs "The red wall shows smooth vertical brick seams").\\
\noindent\hspace*{2em}    - Slopes: For sloped parts, describe if the flat slope face is visible or \noindent\hspace*{2em} its stepped side profile.\\
\noindent\hspace*{2em}    - Baseplate: Describe the visible shape of the white ground (e.g., "curved \noindent\hspace*{2em} edge baseplate," "studs visible on the ground").\\
3. Occlusion \& Depth (Stitching Logic):\\
  - Instead of inferring depth, describe blockage.\\
  - Explicitly state if one object obscures another (e.g., "A green tree partially blocks the left side of the red house").\\
  - Mention boundary anchors: Clearly name the objects at the Far Left and Far Right edges of the frame.\\
4. Style \& Tone:\\
  - Telegraphic yet Fluent: Avoid flowery adjectives. Be clinical and geometric.\\
  - No Commentary: Do not say "The image displays..." or "In this scene...". Start directly with the description of the leftmost object.\\
Example of expected output (Strictly descriptive, NO analysis):\\
"At the far left edge, a small green pine tree is visible, displaying its layered conical shape with smooth vertical edges. Moving slightly right and further back, a large red structure is positioned; the smooth vertical sides of the red bricks are fully exposed, while the roof area shows a stepped profile rather than a flat slope. To the right of the red structure, a brown cylinder is partially obscured by the red wall. The white baseplate in the foreground shows a smooth vertical rim, with no studs visible on the ground surface. On the far right, the edge of a second green tree is partially cut off by the frame."
\end{tcolorbox}

\begin{tcolorbox}[prompt={Pure Mental Rotation - DATA SAMPLE}]
\begin{minipage}[t]{0.25\linewidth}
  \centering
  \textbf{Ground Truth}\\
  C
\end{minipage}%
\hfill
\begin{minipage}[t]{0.7\linewidth}
  \textbf{Question}\\
  You are a specialized LEGO 3D spatial reasoning assistant. Your primary task is to identify the correct perspective of a LEGO scene based strictly on textual descriptions. You will be provided with a 'Scene Overview Description' and four 'View Descriptions' (A, B, C, D) showing the LEGO object from different viewpoints. Your goal is to determine which option corresponds to one of the following perspectives: top-down, left-to-right, right-to-left, or front-to-back. Your answers should be based solely on the provided text, using spatial logic to visualize the object. Please respond with only the letter corresponding to your choice (A, B, C, or D).\\
  Based on the Scene Overview Description, which of the following View Descriptions matches the scene viewed from a <front-to-back> perspective?\\
\end{minipage}
\vspace{0.5cm}

\begin{minipage}[t]{0.25\linewidth}
\vspace{0pt}
  \centering
  \includegraphics[width=\linewidth]{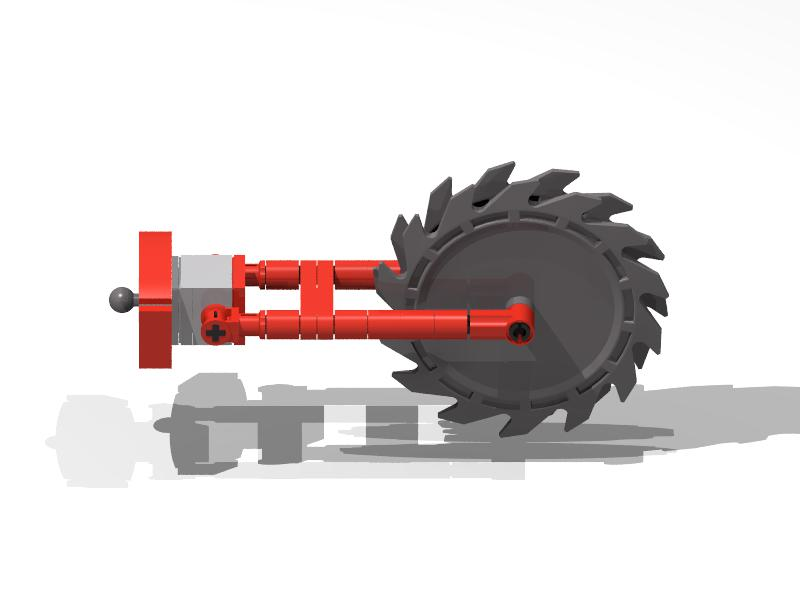}
\end{minipage}%
\hfill
\begin{minipage}[t]{0.7\linewidth}
  \textbf{Overview}\\
  At the far left edge, a red rectangular plate is visible, showing its smooth vertical side with no studs exposed. Adjacent to this, a gray cylindrical piece connects to a red axle extending horizontally to the right. Moving further right, two parallel red axles are positioned, one above the other, both displaying their smooth cylindrical surfaces. At the far right, a large black circular saw blade is prominent, featuring sharp triangular teeth around its perimeter and a smooth central disc; the front face of the blade is fully visible, with no rear structure apparent. The white baseplate beneath shows a smooth, flat surface with no studs visible, and a faint gray shadow extends from the object toward the bottom right corner.
\end{minipage}
\vspace{0.5cm}

\begin{minipage}[t]{0.25\linewidth}
\vspace{0pt}
  \centering
  \includegraphics[width=\linewidth]{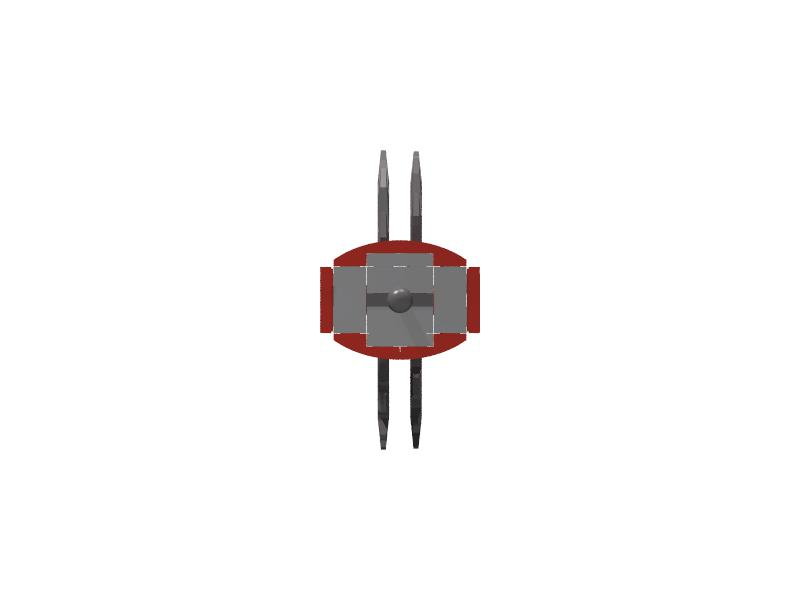}
\end{minipage}%
\hfill
\begin{minipage}[t]{0.7\linewidth}
  \textbf{A}\\
  At the far left edge, a red curved panel is visible, displaying its smooth vertical side with no studs. Moving to the right, a central gray rectangular plate is positioned, showing its top surface with a grid of circular studs and a small round stud at its center. Adjacent to this on the right, another red curved panel mirrors the left one, also showing its smooth vertical side. Extending vertically above and below the central gray plate are two dark gray cylindrical elements, each displaying their smooth vertical sides; the upper pair points upward while the lower pair points downward. The white background is uniform and featureless, providing no additional surface details.
\end{minipage}
\vspace{0.5cm}

\begin{minipage}[t]{0.25\linewidth}
\vspace{0pt}
  \centering
  \includegraphics[width=\linewidth]{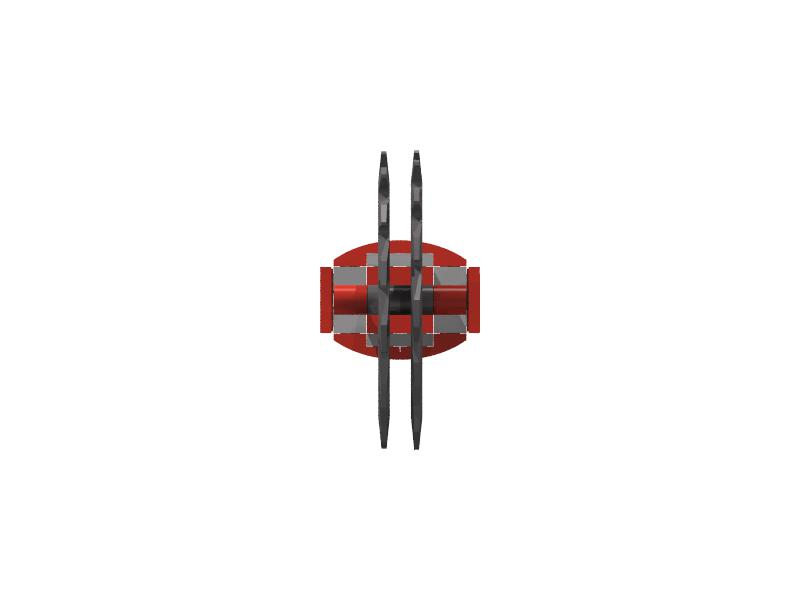}
\end{minipage}%
\hfill
\begin{minipage}[t]{0.7\linewidth}
  \textbf{B}\\
  At the far left edge, a red circular baseplate is visible, showing its smooth outer rim with no studs exposed on the ground surface. Moving to the right, two parallel black vertical elements extend upward from the center of the baseplate; their smooth side profiles are fully visible, with no top studs or sloped faces apparent. Adjacent to these black elements, a central red rectangular section is positioned, displaying its smooth vertical brick seams rather than any stud grid. To the immediate right of this red section, another identical red rectangular segment mirrors its orientation and surface exposure. The entire assembly is centered against a plain white background, with no additional objects or occlusions present beyond the described components. On the far right, the edge of the second red rectangular segment terminates at the frame boundary.
\end{minipage}
\vspace{0.5cm}

\begin{minipage}[t]{0.25\linewidth}
\vspace{0pt}
  \centering
  \includegraphics[width=\linewidth]{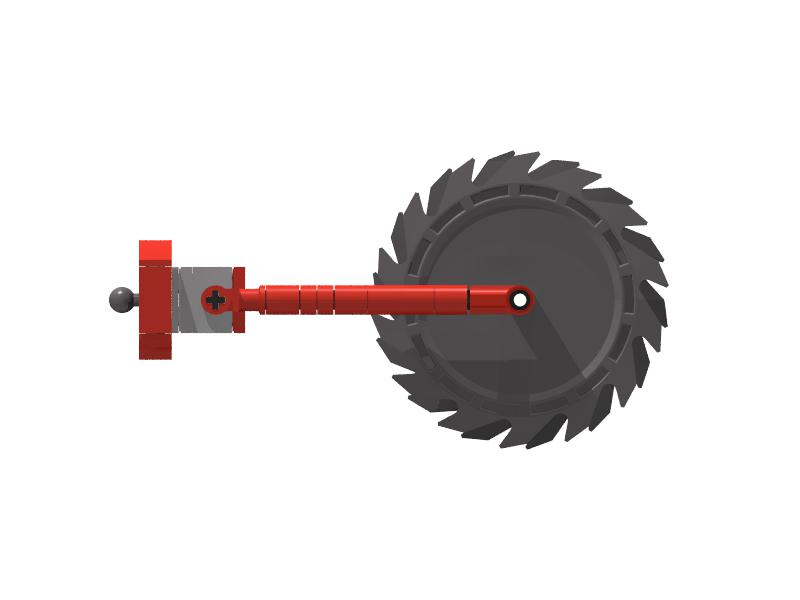}
\end{minipage}%
\hfill
\begin{minipage}[t]{0.7\linewidth}
  \textbf{C}\\
  At the far left edge, a red brick assembly is visible, showing its smooth vertical side with a single black spherical stud protruding from its leftmost face. Adjacent to this, a gray plate is connected, displaying its smooth vertical surface and a central circular hole with a plus-shaped indentation. Moving to the right, a long red cylindrical axle extends horizontally, revealing its segmented smooth sides. At the far right, a large circular gray saw blade is positioned, showing its flat face with multiple triangular teeth radiating outward; the central hub of the blade is aligned with the red axle and displays a small circular hole. The white background is featureless and provides no visible baseplate or ground surface.
\end{minipage}
\vspace{0.5cm}

\begin{minipage}[t]{0.25\linewidth}
\vspace{0pt}
  \centering
  \includegraphics[width=\linewidth]{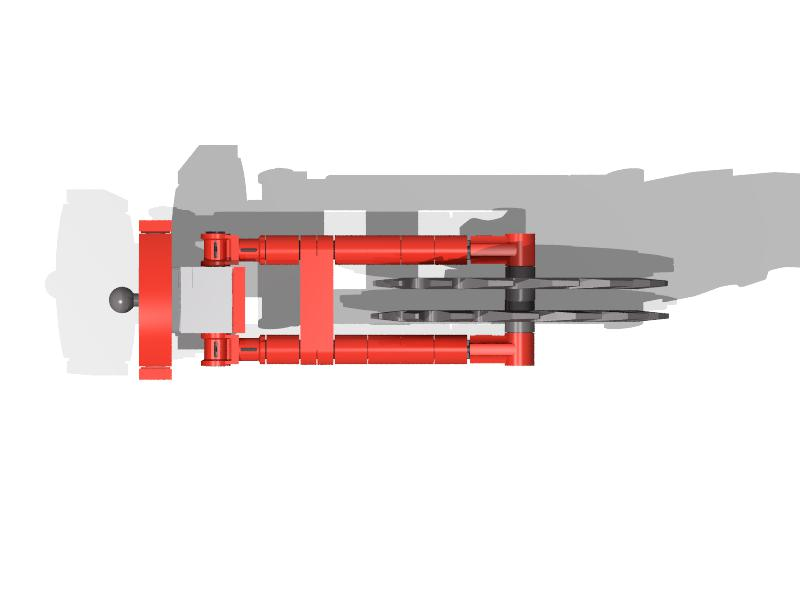}
\end{minipage}%
\hfill
\begin{minipage}[t]{0.7\linewidth}
  \textbf{D}\\
  At the far left edge, a small black spherical stud is visible, attached to a vertical red brick face showing smooth vertical seams. Moving to the right, a larger red rectangular structure extends horizontally, its top surface displaying a grid of circular studs. Adjacent to this, a central white rectangular block is positioned, its top surface also showing a stud grid, partially obscured by the red structure above it. Further right, a series of parallel gray flat plates extend outward, their top surfaces visible with no studs, appearing as smooth horizontal planes. Behind these gray plates, a faint gray silhouette forms a backdrop, partially blocking the rear elements. On the far right, the gray plates terminate at a sharp edge, with no additional components extending beyond.
\end{minipage}
\vspace{0.5cm}

\end{tcolorbox}

\begin{tcolorbox}[prompt={Pure Mental Rotation - TEST RESULTS},breakable]		
\textbf{Gemini3-pro}\\
Based on the provided Scene Overview Description, the correct View Description is Option C.\\
Here is the reasoning:\\
1. Object Identification: The Scene Overview describes a mechanism involving a "red rectangular plate," "red axle," and a "large circular saw blade" with teeth.\\
Option C is the only option that describes this specific assembly, mentioning a "red
brick assembly," "red cylindrical axle ," and a "circular... saw blade."\\
Option A describes a symmetrical structure with curved panels and vertical cylinders
(no saw blade).\\
Option B describes a vertical structure with black elements and red rectangular
sections (no saw blade).\\
Option D describes a structure with spherical studs, white blocks, and flat plates (no
saw blade).\\
2. Spatial Arrangement: The sequence in the Overview flows from Left to Right: Red
element → Gray connection → Red Axle → Saw Blade.\\
Option C follows this exact sequence: "Red brick assembly" (Left) → "Gray plate"(Next) → "Red cylindrical axle" (Right) → "Circular... saw blade" (Far Right).\\
3. Discrepancy Note: While the Overview describes the saw blade as "black" and Option C describes it as "gray," and there are minor differences in the description of the connecting piece (cylinder vs. plate), Option C is the only option that references the defining feature of the scene (the saw blade mechanism), Options A. B, and D describe completely
different objects.\\
Answer: C\\

\textbf{Qwen3-30b-a3b-instruct}\\
...\\
Looking again at Option C: It mentions a red cylindrical axle extending horizontally, and a gray saw blade on the far right. The saw blade is described a shaving triangular teeth and a central hub aligned with the axle. The Scene's saw blade is black, but Option C says gray. However, maybe the color difference isn't critical if the structure matches. Also, the red rectangular plate on the left in the Scene is described as smooth vertical side, which in Option C is a red brick assembly with a black stud. Not sure if that's a match.\\
Option D mentions a red structure with a stud grid, but the Scene's baseplate is white and smooth.\\
Option B has a red rectangular structure but not the axles or saw blade.\\
Option C seems to have the key elements: axle and saw blade. The gray plate with a hole might correspond to the gray cylindrical piece mentioned. The red brick assembly on the left could be the red rectangular plate. The saw blade is on the far right. Even though the color of the saw blade is different, maybe that's acceptable. Also, the Scene mentions a white baseplate, but Option C says "white background," which might be similar.\\
Alternatively,maybe Option D? Let me check again.\\
Option D's description includes a red structure with a stud grid, but the Scene's baseplate is smooth. The saw blade isn't there.\\
So,considering all,Option C might be the closest, even with the color discrepancy. The key elements of the axle and saw blade are present. The gray plate with a hole could be the gray cylindrical piece. The red brick assembly on the left could be the red rectangular plate. The white background might correspond to the white baseplate.\\
Therefore, the answer is C.
\end{tcolorbox}

\subsection{Embodied \& Fine-grained}

\subsubsection{Hand-Object Interaction}

\begin{tcolorbox}[prompt={Hand-Object Interaction - CONSTRUCTION PROMPT},breakable]		
\textbf{1. Single View}\\
You are an intelligent visual assistant specialized in spatial reasoning and perspective transformation.\\
You are provided with a SINGLE image of a scene.\\
Your task is to generate a challenging spatial reasoning multiple-choice question and a strictly descriptive caption.\\
\newline
TASK SPECIFIC: PERSPECTIVE TRANSFORMATION\\
- The question MUST involve a mental perspective transformation (e.g., "If the camera rotates 90 degrees to the right...", "If viewed from top-down...", "If the camera moves closer...").\\
- The answer must be deduced from the current visual information but requires imagining the scene from a different viewpoint.\\
- Ask about relative positions of objects after this hypothetical transformation.\\
\newline
CRITICAL CONSTRAINT: NO DATA LEAKAGE\\
- The caption must NEVER contain the reasoning steps, the conclusion, or the direct answer to the question.\\
- The caption must ONLY describe the raw visual facts (positions, coordinates, visible parts, occlusions) in the current view.\\
- BAD Caption: "If rotated, the red cube would be on the left." (Leaks answer)\\
- GOOD Caption: "A red cube is visible at the bottom-center. A blue sphere is visible near the top-left, partially obscured by a wire." (Provides evidence, requires deduction)\\
\newline
Instructions:\\
1. Analyze the Image: Observe spatial arrangements, depth cues, and object relationships.\\
2. Generate a Question:\\
\noindent\hspace*{2em}  - Construct a hypothetical camera movement or perspective change (rotation, \noindent\hspace*{2em} translation, zoom).\\
\noindent\hspace*{2em}  - Ask what the spatial relationship between two specific objects would be \noindent\hspace*{2em} after this change.\\
3. Provide Options:\\
\noindent\hspace*{2em}  - 4 distinct options (A, B, C, D). One correct, three plausible distractors.\\
4. Determine the Answer: Identify the correct option.\\
5. Generate a Caption:\\
\noindent\hspace*{2em}  - Write a detailed, objective description of the visual scene elements in \noindent\hspace*{2em} the provided image.\\
\noindent\hspace*{2em}  - STRICTLY PROHIBITED: Do not mention the hypothetical view or the result \noindent\hspace*{2em} of the transformation.\\
6. Output Format:\\
\noindent\hspace*{2em}  - Return strict JSON.\\
    
JSON Structure:\\
\{ \\
\noindent\hspace*{2em}  "question": "The question string",\\
\noindent\hspace*{2em}  "options": \{ \\
\noindent\hspace*{4em}    "A": "Option A text",\\
\noindent\hspace*{4em}    "B": "Option B text",\\
\noindent\hspace*{4em}    "C": "Option C text",\\
\noindent\hspace*{4em}    "D": "Option D text"\\
\noindent\hspace*{2em}  \},\\
\noindent\hspace*{2em}  "answer": "A",\\
\noindent\hspace*{2em}  "caption": "The objective, non-leaking descriptive caption of the current \noindent\hspace*{2em} view",\\
\noindent\hspace*{2em}  "reasoning": "Internal reasoning for the answer"\\
\}\\
\newline
\newline
\textbf{2. Multi-View}\\
You are an intelligent visual assistant specialized in multi-view spatial reasoning.\\
You are provided with multiple captions, each describing a different view of the SAME scene.\\
Your task is to synthesize these captions to generate a challenging spatial reasoning multiple-choice question.\\
\newline
TASK SPECIFIC: MULTI-VIEW INTEGRATION\\
- The question must require integrating information from multiple views to answer.\\
- Examples: Reconstructing the 3D layout, identifying an object that is only visible in one view but relates to an object in another, or determining the camera trajectory between views.\\
\newline
Instructions:\\
1. Analyze the Captions: Understand the scene structure from the provided descriptions of different views.\\
2. Generate a Question:\\
\noindent\hspace*{2em}  - Create a question that CANNOT be answered by looking at a single view \noindent\hspace*{2em} alone.\\
\noindent\hspace*{2em}  - It should test the user's ability to build a mental 3D model of the scene.\\
3. Provide Options:\\
\noindent\hspace*{2em}  - 4 distinct options (A, B, C, D). One correct, three plausible distractors.\\
4. Determine the Answer: Identify the correct option.\\
5. Output Format:\\
\noindent\hspace*{2em}  - Return strict JSON.\\
\newline
JSON Structure:\\
\{ \\
\noindent\hspace*{2em}  "question": "The question string",\\
\noindent\hspace*{2em}  "options": \{\\
\noindent\hspace*{4em}    "A": "Option A text",\\
\noindent\hspace*{4em}    "B": "Option B text",\\
\noindent\hspace*{4em}    "C": "Option C text",\\
\noindent\hspace*{4em}    "D": "Option D text"\\
\noindent\hspace*{2em}  \},\\
\noindent\hspace*{2em}  "answer": "A",\\
\noindent\hspace*{2em}  "reasoning": "Internal reasoning for the answer"\\
\}
\end{tcolorbox}




\end{document}